\documentclass[lettersize,journal]{IEEEtran}
\usepackage{amsmath,amsfonts}
\usepackage{algorithmic}
\usepackage{algorithm}
\usepackage{array}
\usepackage[caption=false,font=normalsize,labelfont=sf,textfont=sf]{subfig}
\usepackage{textcomp}
\usepackage{stfloats}
\usepackage{url}
\usepackage{verbatim}
\usepackage{graphicx}
\usepackage{cite}
\usepackage{booktabs}
\usepackage{xspace}%
\newcommand{\fig}{Fig.\xspace}

\newcommand{\tbl}{Table\xspace}
\newcommand{\eg}{e.g.,\xspace}

\newcommand{\ie}{i.e.,\xspace}
\newcommand{\sect}{Section\xspace}
\newcommand{\quotes}[1]{``{#1}''}
\newcommand{\quoquo}[1]{`{#1}'}

\begin{document}

\title{Human-Robot Interaction Conversational User Enjoyment Scale (HRI CUES)} 

\author{Bahar Irfan
, Jura Miniota
, Sofia Thunberg
, Erik Lagerstedt
, Sanna Kuoppamäki
, Gabriel Skantze
, Andr\'{e} Pereira
\thanks{Bahar Irfan, Jura Miniota, Gabriel Skantze, and Andr\'{e} Pereira are with the Division of Speech, Music and Hearing at the KTH Royal Institute of Technology, 100 44 Stockholm, Sweden. E-mail: \{birfan, jura, skantze, atap\}@kth.se.}
\thanks{Sofia Thunberg is with the Department of Computer Science and Engineering, at Chalmers University of Technology and Gothenburg University, 412 96 Gothenburg, Sweden. Email: sofia.thunberg@chalmers.se.}
\thanks{Erik Lagerstedt is with the Department of Philosophy, Linguistics, Theory of Science at the University of Gothenburg, 405 30 Gothenburg, Sweden. Email: erik.lagerstedt@gu.se.}
\thanks{Sanna Kuoppamäki is with the Division of Health Informatics and Logistics at the KTH Royal Institute of Technology, 141 57 Huddinge, Sweden. E-mail: sannaku@kth.se.}
\thanks{This work was supported by KTH Digital Futures (Sweden) and the Swedish Research Council project 2021-05803.}} 

\markboth{IEEE Transactions on Affective Computing,~Vol.~, No.~, ~}
{}


\maketitle

\begin{abstract}

    Understanding user enjoyment is crucial in human-robot interaction (HRI), as it can impact interaction quality and influence user acceptance and long-term engagement with robots, particularly in the context of conversations with social robots. However, current assessment methods rely solely on self-reported questionnaires, failing to capture interaction dynamics. This work introduces the Human-Robot Interaction Conversational User Enjoyment Scale (HRI CUES), a novel 5-point scale to assess user enjoyment from an external perspective (\eg by an annotator) for conversations with a robot. The scale was developed through rigorous evaluations and discussions among three annotators with relevant expertise, using open-domain conversations with a companion robot that was powered by a large language model, and was applied to each conversation exchange (\ie a robot–participant turn pair) alongside overall interaction. It was evaluated on 25 older adults' interactions with the companion robot, corresponding to 174 minutes of data, showing moderate to good alignment between annotators. Although the scale was developed and tested in the context of older adult interactions with a robot, its basis in general and non-task-specific indicators of enjoyment supports its broader applicability. The study further offers insights into understanding the nuances and challenges of assessing user enjoyment in robot interactions, and provides guidelines on applying the scale to other domains and populations. The dataset is available online\footnote{HRI CUES Dataset (anonymized transcripts, annotation scores, and self-reported user perceptions): \url{https://doi.org/10.5281/zenodo.12588810}}. 
\end{abstract}


\section{Introduction}\label{intro}

User enjoyment, referring to the user's subjective perception and experience of the 
enjoyment of interaction, 
is an important indicator of acceptance of robots and willingness to engage with them over time~\cite{heerinkElderlyEnjoyment}. Particularly in the context of conversational agents or companion robots, where the primary goal often revolves around providing emotional support or companionship, enjoyment serves as a vital metric for evaluating the effectiveness of such systems. Therefore, developing reliable and efficient methods to measure user enjoyment in Human-Robot Interaction (HRI) scenarios is essential for designing and improving future generations of robots.


\begin{figure}[t]
    \centering
    \includegraphics[width=\columnwidth]{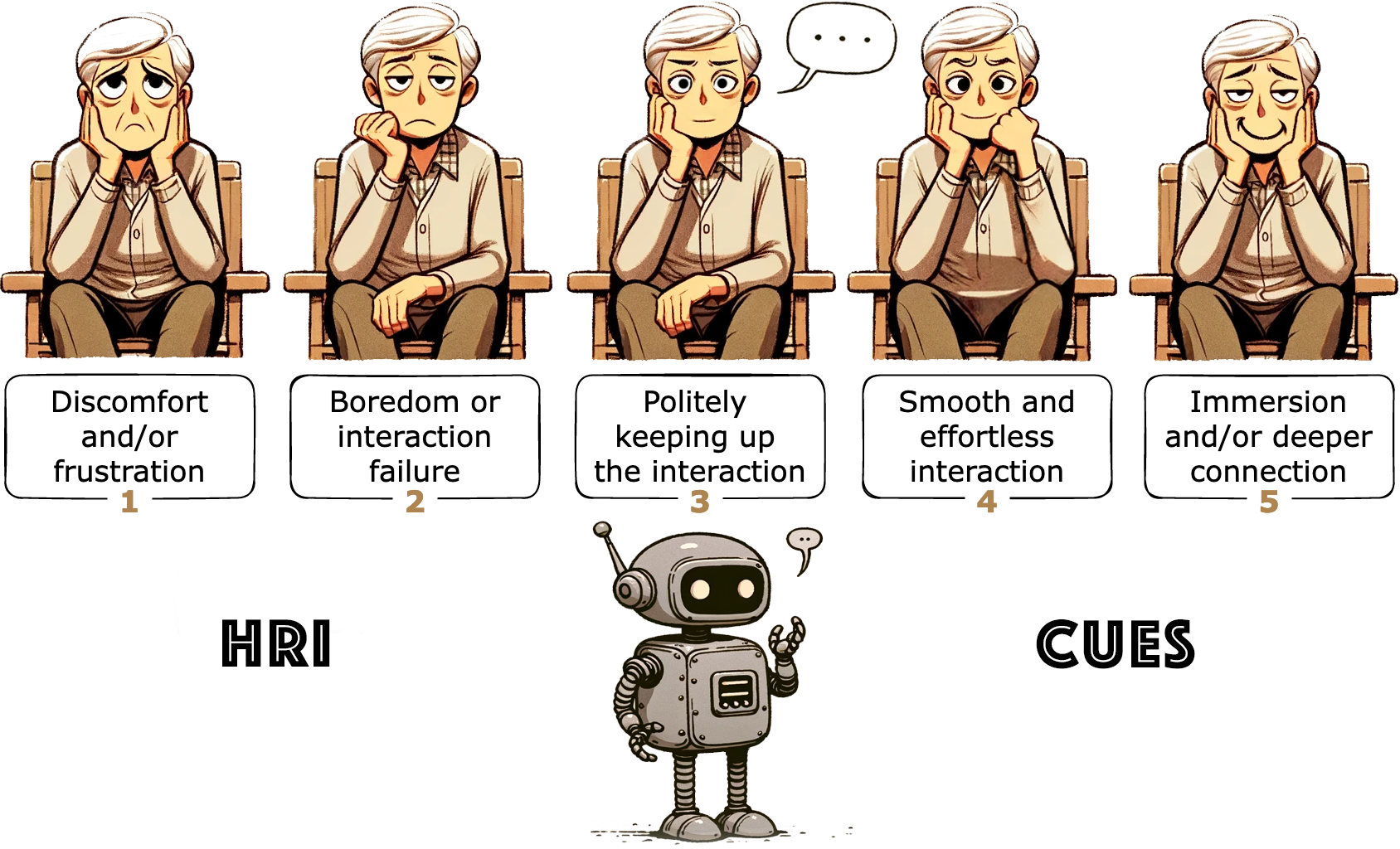}
    \caption{Human-Robot Interaction Conversational User Enjoyment Scale (HRI CUES).} 
    \label{fig:HRI-CUES}
\end{figure}

User enjoyment is closely linked to the intention to use robots, particularly among older adults~\cite{heerinkElderlyEnjoyment}. Conversational companion robots are often developed to provide social or emotional support to older adults in a home or care home environment~\cite{Abdi2018}. 
Prior studies in HRI explored older adults' acceptance, use, and interaction with robots, showing that 
older adults experience difficulties in interacting with a conversational agent, 
such as in hearing and understanding the system, and responding to it~\cite{kuoppamaki_designing_2023}, consequently attributing a low level of social acceptance to the robot~\cite{Thunberg2022}. 
Some of these challenges emerged from the limitations of earlier systems and approaches, which relied heavily on scripted and non-adaptive interactions, underscoring the need for more autonomous and socially capable robots. 
The recent introduction of Large Language Models (LLMs) has enabled the development of companion robots equipped with social capabilities, eliminating the need for scripted interactions or Wizard of Oz, which was the common approach in conversational HRI studies (\eg~\cite{kuoppamaki_designing_2023, Thunberg2022}), and the inherent human influence that hinders the construction of robots capable of autonomously mitigating errors~\cite{breazeal2005effects, riek2012wizard}. Recent studies applied LLMs to conversational robots in various domains, including therapy~\cite{lee2023developing}, service~\cite{cherakara2023FurChat}, and care for older adults~\cite{khoo2023spill, irfan2025between}, which demonstrate their potential and limitations in diverse contexts that lead to enjoyable or unpleasant experiences, further showing the importance of detecting user enjoyment during conversations with robots.
 

Sustaining enjoyment, especially in daily encounters such as for companion robots, is a challenging task yet to be solved. User engagement and satisfaction may fluctuate within day-to-day interactions, but also within the interaction itself, based on the robot's performance in conversation flow, content, and contextual memory, which may affect user enjoyment. However, relying solely on user engagement or task satisfaction as an indicator of enjoyment can be misleading, as these are distinct concepts that may occur simultaneously but can also be mutually exclusive, that is, one may exist without the other depending on the context of the interaction. For instance, a user can be engaged in an argument or be satisfied with the agent completing a task, but not necessarily enjoy either interaction. 
Despite several studies detecting user engagement~\cite{oertel2020engagement} or satisfaction~\cite{Higashinaka2010IssuesIP,Wei2021multimodal} with automatic measures, 
measure of user enjoyment is limited to self-reports from users~\cite{perrig2024measurement, kono2022fun}. Not only can self-reports be unreliable due to demand characteristics, self-presentation, or the Hawthorne effect (stemming from conformity to perceived norms or researcher expectations~\cite{irfan2018social}), but they represent overall feedback of the interaction rather than a continuous, moment-by-moment measure. 
While affect recognition systems can detect laughter and smiles~\cite{lingenfelser2014event}, 
enjoyment is a complex feeling that can be conveyed through other multimodal cues (see \sect~\ref{scale}). Even in interpersonal communication, enjoyment has been analyzed from an external perspective mainly within the context of marriage~\cite{reimnitz2022enjoyment} or computer entertainment~\cite{jegers2007pervasive}. Thus, there is no scale or an automatic system for assessing user enjoyment in conversations with a robot, the former being required to develop the latter.

This work contributes with the Human-Robot Interaction Conversational User Enjoyment Scale (HRI CUES), illustrated in \fig~\ref{fig:HRI-CUES}, which is a novel 5-point scale for assessing enjoyment in conversations with robots from an external (third-party) perspective. The scale was developed through rigorous annotator discussions based on individual open-domain conversations of three older adults with a companion robot using an LLM, and was tested on 25 older adults' conversations with the same robot, achieving moderate to good annotator alignment. This 1-item scale is not intended to be psychometric; rather, it serves as a methodological tool for annotators to assess user enjoyment construct in HRI, by considering fine-grained conversation exchange levels (\ie turn-by-turn) and the overall interaction level. The scale also holds potential for autonomously identifying enjoyment in real-time HRI~\cite{pereira2024multimodal,janssens2025online}. HRI CUES is presented in a generic form, which has to be adapted to each specific use case (as described in \sect~\ref{sec:otherdomains}). In the particular case described in this paper, open-ended conversations with older adults in Sweden were used (described in \sect~\ref{sec:use-case}) both for the empirical backbone in the development of HRI CUES and for proof of concept. In addition, by providing a detailed exploration of the instances of annotators' concordance and divergence, based on turn-by-turn analysis of enjoyment, in addition to the underlying reasons for discrepancies between users' 
self-reported enjoyment ratings based on metrics typically used in HRI studies, the study offers invaluable insights for understanding the nuances and challenges of assessing user enjoyment in interactions with robots. 

\section{Background and Related Work}\label{background}
\subsection{Defining enjoyment} \label{sec:assessEnjoyment}
Evaluating the user's subjective experience is extremely important in Human-Computer Interaction (HCI) and HRI, with \quoquo{enjoyment} being a frequently assessed metric, e.g.,~\cite{heerinkElderlyEnjoyment, venkatesh2012UTAUT, piasek2018acceptance, kono2022fun}. In this paper, we conceptualize enjoyment as \textit{a felt experience whose nature is primarily determined by two key characteristics: first, its hedonic quality, indicating the extent to which it is perceived as pleasant or unpleasant; and second, its intensity level, describing how activated or subdued the experience is}. This approach allows us to consider not only intense positive states like \quoquo{flow} and relaxed positive feelings, but also to acknowledge and measure experiences involving negative hedonic qualities. Such a comprehensive view of enjoyment is vital for a holistic assessment of user affect in nuanced interactions.

A previous common definition of enjoyment is being in the state of \textit{flow}~\cite{csikszentmihalyi2014flow}. Flow is defined as the optimal experience, which provides a deep sense of enjoyment. It happens when an individual is fully engaged in a task that provides an optimal amount of challenge. Flow is characterized by a set of factors, such as a fading sense of \quoquo{self}, 
a sense that duration is altered, and deep and effortless involvement in the task. The theory states that enjoyment is not obtained in a relaxed state, that it is necessary to be challenged, and links repeated experiences of flow to mastery of a skill.

Another theory derives from the flow theory to define \textit{true fun} as the experience that occurs when a person is experiencing flow, playfulness, and connection all at the same time~\cite{price2021power}. If one or two of the three components that constitute true fun are present, the experience will make a person feel joy or satisfaction, but not true fun. Similarly to what characterizes being in a state of flow, experiencing true fun is characterized by losing track of time, letting go, and being completely present in the moment, with the addition of laughter, feeling free, a sense of child-like excitement, and joy.

User enjoyment can align with engagement as evidenced in the theory of flow where a sense of deep focus is highlighted. However, a person might be deeply engaged in an argument, being focused, responsive, and mentally invested, yet not enjoy the interaction at all~\cite{jolly2019wanting}. Conversely, someone might enjoy the atmosphere of a group conversation without being particularly engaged with the robot or any individual speaker. Similarly, user satisfaction is related to enjoyment and is defined as \quotes{freedom from discomfort and positive attitudes towards the use of the product}~\cite{iso9241}. However, user satisfaction is an evaluation of the user's experience of a system performing a task, focusing on the task itself. For instance, a user might report satisfaction with a robot's performance because it met their expectations or completed a task successfully, even if the interaction itself was not enjoyable. 

Other theories recognize enjoyment in less intense states without focusing on the difficulty of the task. For instance, flow-like states can be differentiated from the overall positive valence of an experience~\cite{mekler2014systematic}. This relates to the circumplex model of emotion~\cite{russell1980circumplex}, which features arousal (low to high) on one axis and pleasure or valence (negative to positive) on the other axis~\cite{ekman1992argument}. Certain theories of enjoyment focus on high valence values that can contain lower-engagement positive emotions (\eg content or calm), while others prioritize high-arousal states (\eg excitement)~\cite{mekler2014systematic}. The \quoquo{happy} emotion in the circumplex model reflects a balance of high arousal and positive valence.

Our work aims to capture both higher levels of arousal and, especially, valence, while incorporating other elements for classifying lower levels of enjoyment. Casual, open-domain conversations between older adults and robots may involve aspects of both arousal and a range of emotions. As such, both dimensions are relevant to our holistic model of enjoyment.

\subsection{Assessing enjoyment} 
Enjoyment is generally evaluated through self-reported questionnaires tailored to the specific application domain. For instance, the Quality of Life Enjoyment and Satisfaction Questionnaire (Q-LES-Q)~\cite{endicott1993quality} and Physical Activity Enjoyment Scale (PACES)~\cite{kendzierski1991physical} are used in healthcare applications. In HCI and HRI research, enjoyment is not often the primary focus for evaluating user perceptions, but is typically included as part of a more comprehensive model~\cite{kono2022fun}. Enjoyment frequently appears as a self-reported measure, either as a construct within established models like the Unified Theory of Acceptance and Use of Technology (UTAUT)~\cite{venkatesh2012UTAUT}, or as single-item measures in custom questionnaires (e.g.,\quotes{Did you feel fun?}~\cite{nishimura2021vocal}, \quotes{Was playing with the robot enjoyable/not enjoyable?}~\cite{cooney2011interaction}). 
Enjoyment was found to be highly correlated 
with \quoquo{satisfying}, \quoquo{entertaining}, \quoquo{exciting},\quoquo{fun}, and \quoquo{interesting} in HRI~\cite{lee2006}. Technology Acceptance Model (TAM)~\cite{TAM} was also adapted to HRI by incorporating measures of affect and cognition to improve its accuracy in explaining technology adoption, and this adaptation included questions about perceived enjoyment~\cite{van2019trust}.

User enjoyment has been shown to correlate with the intention to use a robot among older adults~\cite{heerinkElderlyEnjoyment}, highlighting its importance for long-term interactions. The Almere model~\cite{heerink2010almere} is an extended version of the UTAUT that is widely used in research on robots for older adults~\cite{piasek2018acceptance}. It incorporates enjoyment, social interaction, and social influence as factors mediating the acceptance and intention to use robots.

While user enjoyment is commonly measured through self-reporting in HRI, using it as a self-report metric has several limitations, such as conforming to perceived norms or researcher expectations or the (in)ability to recall the events and report correctly from memory~\cite{problems, irfan2018social}. In addition, it is often desirable to estimate what a user is feeling by assessing it from an external perspective when self-reporting is not possible or the goal is to automate behavior at the dialogue exchange level during interactions in real-time. External assessments and self-reports are not mutually exclusive; rather, they can complement each other, with the results of one method potentially serving to validate the other. While prior research has relied on smiles and laughter to automatically classify user enjoyment (\eg in storytelling contexts)~\cite{lingenfelser2014event}, these signals can be ambiguous, contradictory, and highly context-dependent~\cite{ginzburg2015understanding, haakana2010laughter}. Moreover, automatic classification of enjoyment would benefit from incorporating a broader range of multimodal cues, as enjoyment is often expressed through more than just facial expressions.

\subsection{Assessing enjoyment in conversations}
User satisfaction is often used in conversation research~\cite{lee2017enhancing, lee2023understanding, Wei2021multimodal}, but is typically evaluated in relation to a task as previously described in \sect~\ref{sec:assessEnjoyment}~\cite{WALKER1998317, SCHMITT201512}. For instance, the Paradigm for Dialogue System Evaluation (PARADISE)~\cite{WALKER1998317} is a framework for evaluating user satisfaction in dialogue systems, based on self-reported satisfaction on a dialogue level and is influenced by other metrics, such as task success in travel booking and accessing emails. Similarly, \textit{interaction quality}, which evaluates user satisfaction from an external perspective on the exchange level, was analyzed by three annotators in bus schedule inquiries with chatbots over phone calls from a data corpus of 200 dialogues and a lab study with 38 subjects~\cite{SCHMITT201512}. An autonomous system was developed based on their ratings, which correlated highly with them, but not with users' self-reported satisfaction scores, which was attributed to the subjectivity of the measure and variability in user perceptions.
User satisfaction has also been measured in the text domain with 
an annotation protocol similar to our study, based on a dataset of 1000 dialogues between 50 users and a chatbot on attentive listening and conversations about animals~\cite{Higashinaka2010IssuesIP}. Two annotators were recruited and an annotator instruction session was conducted. The annotators were requested to go through the conversation exchanges once, without going back or looking at the history. The annotators used three metrics of user satisfaction: \quoquo{smoothness of the conversation}, \quoquo{closeness perceived by the user towards the system}, and \quoquo{willingness to continue the conversation}, rated from 1 to 7. However, \textbf{no agreement} was found between the annotators, even after changing the granularity of the scale to two levels, low and high, showing the complexity of evaluating a subjective measure from an external perspective. 
Another study by Wei et al~\cite{Wei2021multimodal} developed a multimodal model for user satisfaction based on conversations with a virtual agent. User satisfaction was annotated by the wizard controlling the agent, the users themselves, and external annotators who rated dialogue exchanges on metrics like topic continuance, external sentiment, and self-sentiment. Notably, two metrics directly linked satisfaction to enjoyment. The resulting multimodal model outperformed human annotators in evaluating user satisfaction at the overall interaction level.


Our user enjoyment scale was developed based on the work of Reimnitz and Rauer~\cite{reimnitz2022enjoyment}, which is the only scale identified that specifically evaluates user enjoyment in conversations from an external perspective. The study assessed the enjoyment in conversations between 64 married couples and compared that to each spouse's marital happiness. For measuring enjoyment, they developed a scale for observational coding that took into consideration affective signs and the tone of the interaction. The scale ranged from 1 (very low enjoyment) to 7 (very high enjoyment), with 3 as a neutral anchor. Two annotators took into account both affective signs (\eg mutuality of the interaction, tone of voice, consistent mutual gaze, facial expressions, physical touching, body language) and the tone of the interaction (\ie neutral, enthusiastic, and delightful) when rating enjoyment. The annotators had good to moderate agreement, with intraclass correlation (ICC) on 20\% of the interactions. The study found that couples who displayed high enjoyment in their conversations also reported having a happier marriage. This aligns with prior research in human relationships, which found that mutually enjoyable behavior leads to increased intimacy, trust, security, and satisfaction in long-term relationships~\cite{campbell2008observational}, signaling that enjoyment could be highly influential in achieving long-term HRI.

Similar to this scale, our study analyzes open-domain, dyadic conversations outside of task-oriented settings, but with a focus on evaluating enjoyment in conversations with autonomous robots. Although prior research assessed autonomous conversational systems with related metrics that touch upon aspects of enjoyment, these metrics capture it only indirectly. Our study, instead, emphasizes understanding enjoyment within conversations with robots. To the best of our knowledge, no prior study measures enjoyment in such settings from an external perspective that takes into account multimodal aspects of HRI. Our work seeks to bridge this gap by proposing a scale that captures user enjoyment within conversation at both the exchange and overall interaction levels, with the potential to be used for autonomous systems to adapt the conversations on the fly.

\section{Data}\label{sec:use-case}

As outlined in \sect~\ref{intro}, achieving and maintaining user enjoyment is important for encouraging continued interactions with robots, especially in daily encounters. This becomes particularly prominent for companion robots for older adults that aim to provide social and emotional support to mitigate loneliness in their daily lives. 
This work builds upon the data from~\cite{irfan2025between} on the participatory design development (with two studies) of an autonomous companion robot that integrates an LLM for conversations with older adults, to build the user enjoyment scale for conversational HRI and evaluate enjoyment. 

This section summarizes the robot architecture used in both studies, outlines the studies, and describes how the data was selected from those studies to create the scale (alignment) and evaluate it.


\subsection{Robot Architecture}\label{data:robot}

The Furhat robot was employed in the studies, featuring a neutral-looking face that underwent user validation before interactions. The robot's face engine incorporated smiles and eyebrow raises during conversations to enhance naturalness and provide non-verbal feedback to users without context analysis. To further refine the interactions, the robot incorporated subtle behaviors like blinking, eye shifts, and brief gaze aversion while speaking, based on silences in user input. 

GPT-3.5 (OpenAI, initial study: text-davinci-002, second study: text-davinci-003) was used for dialogue generation, as it was the most capable LLM at the time (September 2022 / March 2023). Prompting was used to give a persona to the robot. Initially, this focused on getting to know more about the conversation partner, however, this created superficial conversations that were not interesting for older adults. That is why, the persona was changed for the latter study to be empathetic, guiding it to ask open questions, listen actively with follow-up questions, and reflect on situations. 

Initially, English was chosen as the communication language for speech recognition (Google Cloud Speech-to-Text), dialogue generation, and synthesis (Amazon Polly) due to the more extensive training data available for LLMs. However, the initial study with Swedish-speaking older adults showed the need for communicating in their native language. Hence, Swedish was used in the follow-up study. 

A USB microphone array (Seeed Studio) was used in both studies to obtain clear audio for speech recognition. A silence-based threshold was used for turn-taking, where the robot would take the turn (generate a response and synthesize it) after a fixed period of silence following the end of the user's speech, without listening to the user while taking the turn. In the initial study, the participants were interrupted frequently. Thus, this threshold was increased for the second study and a red-light signal was added underneath the robot to signal the robot was taking the turn, which substantially decreased the interruptions.

While the conversation with the robot was autonomous, a wizard interface was used to start the interaction with the user by entering the participant ID. The initial and final\footnote{Robot response to start the interaction: \quotes{Hello! I am Furhat, the personalized companion robot. What is your name?} Robot response to end the interaction: \quotes{I would love to talk more another time, but for the sake of time, I need to say goodbye. Thank you for talking with me. Take care!}} robot responses were pre-scripted to ensure that the interaction started and ended the same way for all participants. The rest of the interaction was fully autonomous based on the user's responses and the responses generated by the LLM. The wizard interface was also used to end the interaction if necessary, \ie if the participant wants to end the conversation early or an error occurs in the system that requires a restart to continue the conversation where it is left off. In the initial study, the participant ended the interaction whenever they wanted (by saying \quotes{Goodbye}). In contrast, in the second study, a 7-minute timer was set (checked automatically after each user response), after which the robot would say its pre-scripted response to ensure a fair comparison between users.



\subsection{Study Details}

Ethical approval was granted by the Swedish Ethical Review Authority (reference number 2022-09-21) for the following two studies in~\cite{irfan2025between}. All participants provided informed consent for data recording, analysis, and the use of anonymized (blurred and nameless) images and videos in publications.

\subsubsection{Preliminary Study} 
Preliminary interviews were conducted with Swedish-speaking older adults aged 65 and over, in which they talked with a robot autonomously and individually in English, for 4 to 13 minutes. 
The study had 6 (3 men, 3 women) Swedish-speaking healthy older adults, between 66 to 86 years old ($M=78.3$, $SD=8.3$).\\



\subsubsection{Second Study}\label{data:secondStudy}
Following the preliminary interviews, technical improvements were made for architecture, persona prompt, language, and turn-taking to overcome the interaction failures, as summarized in \sect~\ref{data:robot}. Subsequently, a second study was conducted with 28 older adults having an autonomous open-domain conversation with the robot individually for approximately 7 minutes. Prior to the robot interactions, the robot's capabilities were demonstrated through a researcher having a conversation with the robot (2 minutes), and focus group discussions were made using design scenarios of everyday activities to understand their expectations of companion robots. The researcher(s) were present in the room (to interfere if necessary) during the individual robot interactions. Following the interactions, the participants completed a 68-question Likert scale (1 to 5) questionnaire in~\cite{irfan2025between}, based on HRI (\cite{lee2006, heerink_assessing_2010,  iio2020twin, bartneck2009godspeed, syrdal2009nars, nomura2006anxiety, weiss2009addressing, degraaf2019why}) and open-domain dialogue (\cite{zhang2018personachat, shuster2022blenderbot, borsci_chatbot_2022}), ranging from constructs on user perceptions (\eg enjoyment, ease of use, usefulness, anxiety towards robots) to capabilities of the agent (\eg turn-taking, consistency, fluency of dialogue). For the purposes of this work, only the user enjoyment construct was used, which focuses on user satisfaction, fun, and interestingness of the conversation~\cite{lee2006, heerink_assessing_2010}), also accounting for discomfort in the conversation~\cite{carpinella2017rosas} with the strangeness of the conversation, similar to~\cite{iio2020twin}\footnote{\quotes{Did you feel something strange in that dialogue with the robot?} was used in~\cite{iio2020twin}.}, by reverse-coding it in analysis:
    \begin{enumerate}
        \item I was satisfied with my conversation with the robot.
        \item It was fun talking to the robot. 
        \item The conversation with the robot was interesting.
        \item It felt strange talking to the robot.
    \end{enumerate}
The Cronbach's alpha (for the evaluation data) was $\alpha=0.84$, showing high correlation of items with each other in the construct, with the removal of any item decreasing $\alpha$.

All interactions with the robot were video-recorded by an external camera facing both the participant and the robot at a side angle, as well as through the robot camera to record the participant's face. 

Participants were recruited by distributing the invitation at our university's communication channels, social media, and platforms for gathering senior citizens. In total, 28 (13 men, 15 women) Swedish-speaking healthy older adults between 66 and 86 years old registered as volunteers. We divided this data for the purposes of this study.

\subsection{Alignment Data}\label{data:alignment}

Three videos were chosen for the initial alignment of annotators to 
create HRI CUES (Sections~\ref{familiarization} to \ref{examples-scale}).

Due to the preliminary study being the very first evaluation of an LLM on a social robot with older adults, the interactions had a lot of failures, which were analyzed thoroughly in~\cite{irfan2025between}. Interaction failures can lead to lower likeability and satisfaction~\cite{honig2018understanding} and cause negative tone and emotion in user responses~\cite{kontogiorgos2021systematic}. Thus, to prevent biasing the enjoyment scale solely towards negative experiences, 
but also have an understanding of user reactions to frequent technical failures in current architectures, we chose the most successful interaction (as defined by the least number of failures and the longest length of interaction) from this study to be included as a basis of alignment for developing the enjoyment scale, as described in Sections \ref{familiarization} and \ref{alignment}. The other criterion was that the interaction contained \quoquo{highs and lows}, that is, the participant reacted positively (\eg smile, laugh), neutrally, and negatively (\eg frown, getting impatient) in the video to enable the annotators to understand the spectrum of responses. The corresponding subject (denoted as S0) was an 83-year-old male without prior experience of robots. 
The interaction lasted 13.5 minutes (53 turns). 
The video was recorded from a side angle, facing both the participant and the robot.

In addition, two of the subjects from the second study were selected for annotator alignment: a 69-year-old woman (denoted as S1) and a 75-year-old man (denoted as S2). S1 and S2 did not have any prior interaction with a robot. 
The selection basis was to find interactions containing a range of \quoquo{highs and lows}, as in the previous study, aiming to complement S0 with interactions with fewer failures. S1's interaction lasted 7.5 minutes (27 turns) 
and S2's interaction lasted 7.3 minutes (27 turns). 

Conversational exchanges (\ie a robot turn followed by a participant turn - termed here as a \textit{turn} for brevity) were chosen as the basis of annotations, because they were mostly similar in duration for participants, as well as for paving the way for understanding user enjoyment via autonomous systems to adapt and improve the interaction continuously. As such, the segments to be annotated were created automatically based on the turns in manually-corrected (for timing and content) transcripts. All videos started with the robot's first phrase (greeting of the user). An exchange ends (and a new exchange starts) when the participant stops speaking, as that holds the potential to be detected by an automatic system for evaluating the exchange that could be used to generate a new response. 

\subsection{Evaluation Data}\label{data:evaluation}

The remaining participant interactions from the second study (except one due to lack of a side-video) were used for HRI CUES evaluation (Sections~\ref{annotationEvaluation} and \ref{results}). The resulting data consisted of 25 participants' (12 men, 13 women) interactions, with a mean age of 74.6 ($SD=5.8$). 20 participants had no prior interaction with a robot, and only one had previously talked with a robot. 
Interaction duration was $M=7.4$ minutes ($SD=1.5$) with 12 to 29 turns. Each exchange lasted 5 to 61 seconds ($M=17.7$, $SD=7.2$). 
The total duration of the videos was 174 minutes, corresponding to 590 turns. The videos were segmented using conversational turns, as explained above.

We provide the evaluation dataset for HRI CUES that includes anonymized transcripts, annotation scores, and self-reported user perceptions online~\cite{irfan2024hricuesdataset}\footnote{HRI CUES Dataset~\cite{irfan2024hricuesdataset}: \url{https://doi.org/10.5281/zenodo.12588810}}. Videos of the interactions are available upon request, contingent upon a signed agreement to maintain data confidentiality in accordance with General Data Protection Regulation.

\section{Assessing Enjoyment from Conversations}\label{scaleCreation}

This work addresses the lack of user enjoyment analysis of conversation from an external perspective in HRI. Thus, we started from an existing enjoyment scale in human-human relations to develop HRI CUES, by complementing it with annotations of older adults' interactions with a conversational companion robot. This section describes not only the scale proposed in this work, but also a complete methodology to evaluate enjoyment from videos of conversations with robots. It also provides annotation guidelines and details the practices taken for establishing inter-rater reliability in annotations~\cite{oconnor2020irr}. 


\subsection{Annotator Selection}
Due to the lack of a clear definition of user enjoyment and its subjectivity resulting in high variability in both user perceptions and understanding by a third party, the selection of the right experts as annotators is critical. This is important in general, especially in multidisciplinary fields like HRI, in particular when investigating complex concepts like enjoyment that have relatively different meanings in different academic traditions~\cite{lagerstedt2023multiple}. The annotators in this study should not only be able to detect and understand multimodal cues exhibited by the users to detect enjoyment, but also align well in their perceptions, such that this measure can be used by other researchers based on their understanding and recommendations. In addition, being well-versed in the literature of the user metric in question (user enjoyment) as well as its difference from similar metrics (\eg user satisfaction) is necessary to ensure correlations with prior literature, as well as users' reported perceptions of such metrics.

Familiarity with the target population (participant group) is also important in establishing a better understanding of their needs and reactions. Researchers whose backgrounds focus on HRI with the target population (\eg older or young adults, children, people with disabilities) can put their interactions in context from social, cognitive, and ethnographic perspectives. Annotators also need to be thoroughly familiar with the socio-cultural background of the participants, as culture affects their perceptions of robots and their interactions~\cite{haring2014cultural}. In addition, understanding the nuances and culture-specific idioms (\eg \quoquo{cold turkey}) and proverbs (\eg \quoquo{bite the bullet}) in conversations will be easier for an annotator that is a native speaker.

While a combination of all these aspects is difficult to find in a single annotator, a group of annotators would be able to complement each other, such that during alignment and development of the scale, their horizons can be expanded by the perspectives of the others. While typically two annotators establish inter-rater reliability in qualitative analysis, employing three annotators could better suit the complexity of the task, allowing for tie-breaking and alignment across diverse backgrounds~\cite{koo2016guideline, oconnor2020irr}. Correspondingly, we selected three annotators ($M_{age}= 30$, $SD = 2.94$) who are researchers in the mid-late stages of their PhD, with a background in user enjoyment (Annotator 1, denoted as A1), HRI with older adults and cognitive science (A2), and multimodal HRI and cognitive science (A3). The annotators were native Swedish speakers and thoroughly familiar with Swedish culture.

\subsection{Familiarizing with Data}\label{familiarization}
As a starting point to familiarize annotators with the data and develop a user enjoyment scale for conversational HRI, the annotators were given a slightly adapted version of the enjoyment scale by Reimnitz and Rauer\footnote{1 (very low): no evidence of pleasure. Pair never has fun or enjoys the interaction, although there may be joint interaction. There is no mutual enjoyment of positive affect or negative interaction. 3 (neutral anchor): there is occasional positivity that is not strong or frequently displayed and may be displayed by only one partner towards the other. Pair is doing OK together but without real joy or enthusiasm for their shared interactions. 7 (very high): the pair is very satisfied with the interaction and activity. The couple shows mutual enjoyment in their interaction marked with shared exuberance and/or delight. There is consistent visual regard coupled with affective sharing.\label{marriageScale}}~\cite{reimnitz2022enjoyment} on 
human-human conversations of married couples, 
in which \quoquo{user} was used instead of \quoquo{couple}, and references that relate to couples (\quoquo{mutual enjoyment}, \quoquo{affective sharing}, and \quoquo{exuberance}) were removed. Instead of a 7-point Likert scale, which may be difficult to align given only the lowest, neutral anchor, and highest enjoyment values, a 5-point scale was used: 

\begin{itemize}
    \item 1 (very low): no evidence of pleasure. The user never has fun or enjoys the conversation, although there may be joint interaction.
    \item 3 (neutral anchor): there is occasional positivity that is not strong or frequently displayed. The user does not have real joy or enthusiasm for the conversation.
    \item 5 (very high): the user is very satisfied with the conversation. The user shows enjoyment in their conversation marked with enthusiasm and/or delight.
\end{itemize}
Annotators were encouraged to use the full scale (\ie not abstaining from giving 1 or 5). 

Due to the subjectivity of user enjoyment, it was necessary to establish common grounds on the levels of user enjoyment prior to the annotators analyzing all the robot interactions individually~\cite{oconnor2020irr}. Thus, three exemplar videos (S0, S1, and S2) as explained in \sect~\ref{data:alignment} were chosen that contain a range of negative and positive responses from the user and a variety of technical failures. 

\begin{figure}[t]
    \centering
    \includegraphics[width=0.92\columnwidth]{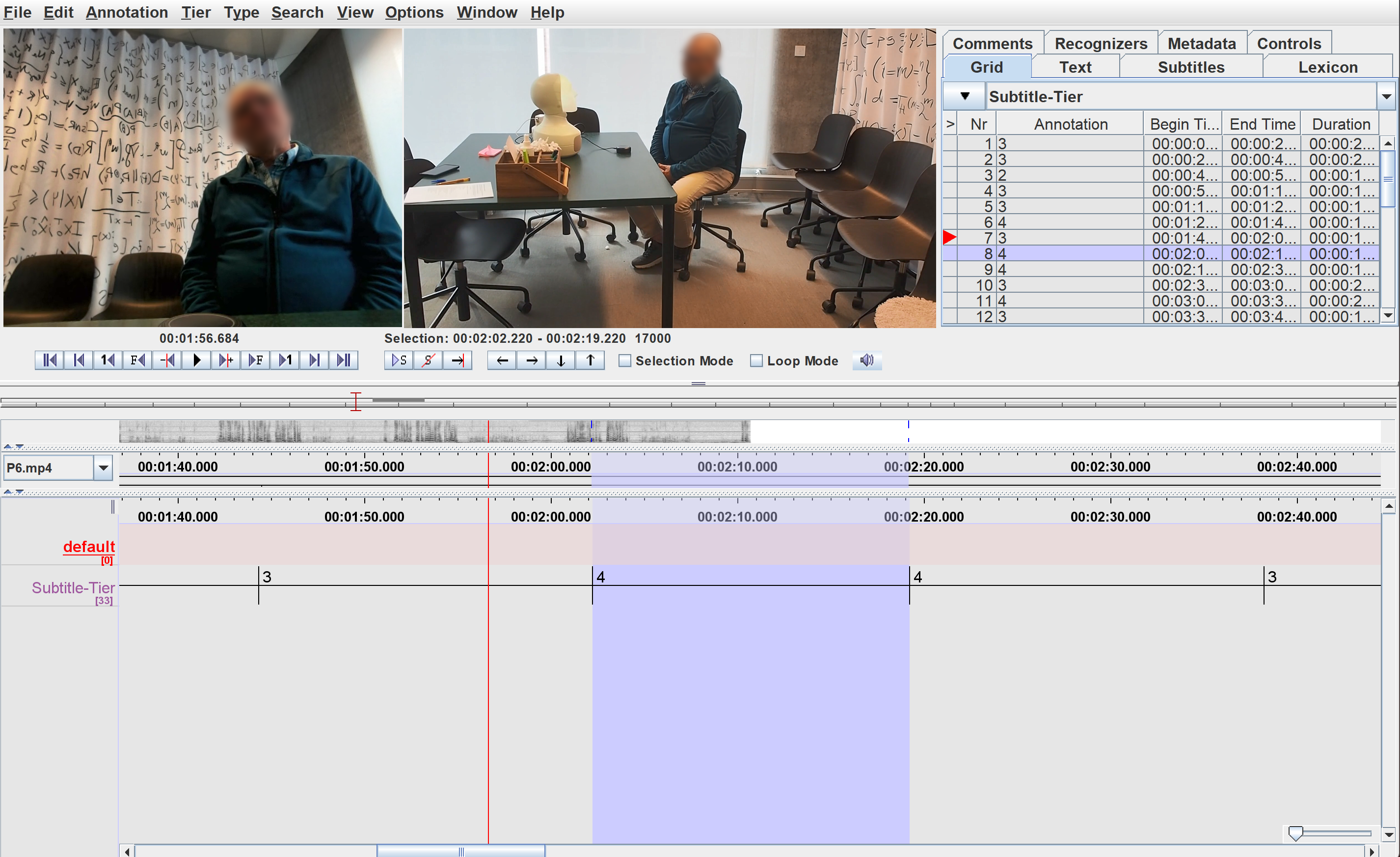}
    \caption{Annotator rating structure of user enjoyment in the ELAN system.} 
    \label{fig:rating}
\end{figure}

In order to incrementally familiarize the annotators with the modalities that a front view (taken from the robot's camera) and side view (external camera facing robot and participant) of robot interaction may introduce, the first exemplar video (S0) contained only the side view, the next one (S1) contained only the front view, and finally the third one (S2) contained both views, as shown in \fig~\ref{fig:rating} and as used in the final annotations. 

Annotators were guided to apply the rating scale on a per-exchange basis, assessing both the robot's response and the participant's subsequent input within their turn. Furthermore, similar to~\cite{Wei2021multimodal}, they were tasked with delivering an overarching assessment of enjoyment encompassing the entire interaction, referred to as \textit{overall enjoyment}. Within this context, annotators were encouraged to provide an in-depth rationale for their ratings, adopting an open-ended approach to offer comprehensive insights. They were asked to elaborate on the aspects and multimodal cues they considered in shaping their evaluations, along with the methodology they employed. Additionally, they were asked to provide whether any challenges or difficulties were encountered while evaluating overall user enjoyment for the interaction and its details. The ratings for each exchange were to be recorded within the ELAN file (\fig~\ref{fig:rating}), while a separate document was designated for annotators to record their overall interaction rating and provide open-ended responses.

\subsection{Annotator Alignment}\label{alignment}

Based on their individual annotations of three robot interaction videos (S0-S2), annotators were asked to meet to align themselves to decide more objectively what each level of the scale corresponds to, such that an agreement can be reached for the analysis of the remaining interaction videos. In addition, they were asked to discuss the aspects and multimodal cues used to give the corresponding scores, in a turn-by-turn fashion, as well as the overall user enjoyment.

To facilitate discussions, a list of aspects and multimodal cues from HRI~\cite{anzalone2015engagement, jung2017affective, skantze2021turntaking, stock2022survey}, HCI~\cite{skantze2021turntaking}, and human-human interaction~\cite{mondada2016challenges, clift_2016, shalihah2018role, rasenberg2022multimodal, reimnitz2022enjoyment} literature was given to the annotators, which were previously used in affective computing, user engagement, user enjoyment, conversation, and turn-taking analysis, in addition to the principal researcher's analysis of the challenges of applying LLMs into conversational robots~\cite{irfan2025between}: 
    \begin{itemize}
        \item \textbf{Facial expressions:} smile, laughter, frown, rolling eyes, sigh, other expressions (\eg smirk, squinting eyes, raising eyebrows). Emotion models~\cite{mehrabian1980pad, russell1980circumplex, occ1988} were described for further context.
        \item \textbf{Gaze:} Mutual gaze, gaze length, gaze aversion, other gaze targets (\eg objects, experimenter) 
        \item \textbf{Body language:} Gestures, gesture duration, gesture frequency, gesture intensity, posture, body orientation, head orientation, arm position (\eg folded/ open), movement, physical contact, pointing, adaptors (\eg touching hair, bouncing legs), nodding/ head shakes,  proximity (distance to the robot) 
        \item \textbf{Vocal features:} Tone, pitch, pace, volume/ loudness, energy 
        \item \textbf{Dialogue responses:} Content, sentiment, length, mirroring, pauses in response, rephrasing/ clarifications, anthropomorphism, disengagement cues (responses that bring the conversation to a halt, \eg \quotes{That is good to know}) 
        \item \textbf{Conversation:} Context, topic, topic initiation, topic closure, topic duration, tone (\eg neutral, enthusiastic, and delightful), vocal fillers (\eg \quotes{uh}, \quotes{erm}), conversation length,  repairs (dealing with failures in interaction), referral to previous topics/parts in a conversation, willingness to talk about personal matters, asking questions about the conversation partner (robot), agreement/ disagreement
        \item \textbf{Turn-taking:} Speaker dominance, willingness to take a turn, interruption, response time, backchannelling 
        \item \textbf{Interacting with others:} Interacting with the experimenter/ third party during the conversation with the robot 
    \end{itemize}

Annotators were encouraged to discuss whether they made use of these elements in their analysis, their usefulness and importance in assessing user enjoyment (even if these aspects were not present in the videos), including any other aspects/ cues they have previously used during familiarization.

Annotators were requested to systematically review the three videos, examining each conversation exchange individually. They were prompted to assess various factors, including rating, cues, and aspects, while also considering the detectability of user enjoyment in each exchange. Annotators were further instructed to identify contrasting and supporting arguments for their ratings, as the reasons behind the divergence between the annotators can be just as, if not more, valuable than the concordance between the annotators~\cite{barbour1115checklists}. Upon completion of the turn-by-turn analysis, annotators were guided to discuss their overall user enjoyment rating, identify which aspects, multimodal cues, and conversation turns contributed most significantly to their conclusions, and consider the relative importance (weights) assigned to these modalities in their assessments. Annotators were also encouraged to evaluate whether specific aspects or cues were observable from both the frontal view (the robot's camera that directly captures participants) and the side view (the external camera that captures both entities) and reflect on how analyzing these different perspectives may have influenced their assessments. Annotators were advised to keep their ratings, both within ELAN and the accompanying document that justifies their scores, readily accessible on their screens as a reference point during the discussions.

Self-reported user perceptions were given to the annotators at this stage, as reported by questionnaire ratings in terms of the level of user satisfaction, fun, interestingness of the conversation, and strangeness of talking to the robot, as described in \sect~\ref{data:secondStudy}. These metrics demonstrate high reliability with each other (Cronbach's $\alpha=0.84$), with lower $\alpha$ when any of the metrics are excluded. The annotators were instructed to view the ratings after the overall user enjoyment in the interaction had been discussed for the corresponding participant. Based on the user perceptions, how these results correlate with their findings and the reasons behind discrepancies were discussed. 

Finally, the annotators were asked to develop a user enjoyment scale for conversational HRI that serves as both a guideline for their remaining annotations and a reference for future research on user enjoyment (\sect~\ref{scale}).

\begin{table*}[]
    \caption{Multimodal signs of disenjoyment and enjoyment for the specific use case reported in section~\ref{sec:use-case}. The signs are presented in no specific order.}
    \centering
    \begin{tabular}{p{.35\textwidth} p{.35\textwidth}}
        \toprule
        \textbf{Signs of disenjoyment} & \textbf{Signs of enjoyment}\\
        \midrule
        Low energy/Tiredness & Smirking \\
        Sighing & Engaged movement (e.g. leaning in) \\
        Long breaths & No strain or discomfort asking questions [to the robot] \\
        Restless movements & Smooth turn-taking \\
        Flat tonality & Dynamic tonality \\
        Silence & Sharing personal experiences [to the robot] \\
        Awkward and negative facial expressions & Sharing an understanding \\
        Flaring nostrils & Anthropomorphizing [the robot] \\
        Disengagement cues & Asking questions to the robot \\
        Topic closure & Being playful \\
        Repeated questions & Flow of conversation \\
        \bottomrule
    \end{tabular}
    \label{tab:signs}
\end{table*}

\subsection{Human-Robot Interaction Conversational User Enjoyment Scale (HRI CUES)}\label{scale}
The discussions between the annotators (A1, A2, and A3) lasted four hours, during which they carefully went through three example videos (S0-S2) that they had previously annotated and discussed the cases one by one. The annotators viewed every exchange in the interaction several times and exchanged the reasons behind their rating in terms of multimodal cues, until they all aligned on the rating for each exchange. Simultaneously, they created a list of signs of enjoyment and dis-enjoyment that they had used during their enjoyment evaluation, which can be seen in Table \ref{tab:signs}. Towards the end of the session, they settled on a 5-item scale based on the initial provided scale, ranking from very low enjoyment to very high enjoyment. The final user enjoyment scale, namely the Human-Robot Interaction Conversational User Enjoyment Scale (HRI CUES), is:
\begin{enumerate}
    \item[1] Very low enjoyment --- Discomfort and/or frustration
    \item[2] Low enjoyment --- Boredom or interaction failure 
    \item[3] Neutral enjoyment --- Politely keeping up the interaction
    \item[4] High enjoyment --- Smooth and effortless interaction
    \item[5] Very high enjoyment --- Immersion 
    in the conversation and/or deeper connection with the robot 
\end{enumerate}


To rate an exchange higher on the user enjoyment scale (4 or 5), the annotators looked for different signs of enjoyment (see Table \ref{tab:signs}), which, for example, were flow of conversation (\ie the topic is moving forward), dynamic tonality, and dynamic phrasing of sentences, as well as sharing an understanding, \ie having common ground with the robot. 

To rate an exchange lower on the scale (1 or 2), the annotators looked for signs of dis-enjoyment (see Table \ref{tab:signs}), which, for example, were restless movements \ie adaptors, such as moving in the chair from side to side or changing arm position. Disengagement cues \eg turning away from the robot, or responding in a way that disrupts the conversation flow, such as \quotes{That is true}), and topic closure (\eg \quotes{Let's talk about something else}) is also an example of disenjoyment. In addition, robot behaviors that disrupted the interaction flow, such as repeated questions, were considered to be strong causes of dis-enjoyment.

Neutral enjoyment (3) refers to a lack of these cues, in which conversation content (and context) becomes more relevant, such as having small talk or continuing the conversation without having much interest in the topic.

In cases where the exchange has cues from multiple scale levels, the annotators determined the dominant level in that interaction. This could be done by observing the intensity of the cues, the significance of the cues, or the interaction trajectory. On the other hand, if an annotator observed strong cues from two moderately or highly distinct levels (as opposed to subsequent levels), they would annotate using a level between those. 
For instance, as evident in this exchange\footnote{Exchange with cues from multiple levels: \url{https://youtu.be/2HA-_5B9JHs}}, when there is discomfort at the beginning (1), but the user continues to politely keep up the interaction (3), the exchange would be annotated as a 2, the mid-point between the levels.

There were also a few cases that were difficult to categorize as enjoyment or dis-enjoyment, and therefore were interpreted as more context-dependent, which, for example, were gaze aversion, attention on the experimenter or camera, topic duration, and initiation. For instance, gaze aversion could be due to thinking, floor management, intimacy regulation (cf.,~\cite{andrist/conversational}), or as a reaction to something the robot said or did.

As general guidelines for annotating user enjoyment, it became clear that it was important to get acquainted with the participants, where different participants had different sets of signals. While watching the videos, the annotators learned each person's rhythm and gestures for what was interpreted as a `baseline' behavior from which the person could deviate during the interaction. This means that the same type of gesture (\eg keeping one's arms crossed) could be interpreted differently for different participants. Instead, an emphasis was placed on the change in behavior. It was also important to separate content from context, \ie it is essential to be mindful of what is being said (conversation content, \eg topic), but the focus should be more on the whole feeling of the exchange. 

The interaction failure does not necessarily refer to a robot failure (\eg incorrect response, speech recognition failure, turn-taking error, disengagement cue), since robot failures can lead to amusement, anthropomorphism, or empathy in the user, thereby increasing user enjoyment. Interaction failure rather refers to the situation when either the user (\eg interrupting the robot) or the robot made a failure that resulted in the interruption being disrupted, leading to low enjoyment.

When annotating the videos, the annotators assumed that in the future, robots would be able to judge the level of user enjoyment in real-time while having a conversation. Therefore, the videos were annotated segment by segment (turn-by-turn), with each segment being watched only once, similar to~\cite{Higashinaka2010IssuesIP}.



\begin{table*}[]
    \caption{Examples for interaction analysis of exchanges for a single participant from the alignment session. Note that enjoyment analysis is dependent on conversation context (\ie previous turns). A more detailed analysis with the corresponding context can be found in \sect~\ref{examples-scale}.}
    \centering
    \begin{tabular}{p{0.05\textwidth} p{0.52\textwidth} p{0.27\textwidth} c}
        \toprule
        \textbf{Turn} & \textbf{Exchange} & \textbf{Reason} & \textbf{Score} \\
        \midrule
        11 & 
        \textbf{ROBOT:} Oh, I just wasn’t sure what you meant. Are you asking why I didn't wanna go with you? \newline
        \textbf{PARTICIPANT:} No, I asked where you want to go. Do you have a favorite place? & 
        Misunderstanding, gaze aversion, sighing, low energy & 
        1 \\
        \addlinespace

        8 & 
        \textbf{ROBOT:} I am sorry, I didn’t understand you. \newline
        \textbf{PARTICIPANT:} Okay. I will have a walk in nature reserve in.. & 
        Interruption, rushed speech, gaze aversion, boredom & 
        2 \\
        \addlinespace

        18 & 
        \textbf{ROBOT:} That sounds great! I love nature too. \newline
        \textbf{PARTICIPANT:} What kind of nature do you like? & 
        Polite continuation, awkward expression & 
        3 \\
        \addlinespace

        23 & 
        \textbf{ROBOT:} Mount Everest is about 8848 meters high. \newline
        \textbf{PARTICIPANT:} Oh, good. Thanks. I didn't know it that exactly. & 
        Smooth interaction, nodding, high energy & 
        4 \\
        \addlinespace

        22 & 
        \textbf{ROBOT:} I can try. \newline
        \textbf{PARTICIPANT:} How high is Mount Everest? & 
        Quick response, smiling, leaning forward, immersion & 
        5 \\
        \bottomrule
    \end{tabular}
    \label{tab:exchange_scores}
\end{table*}

\subsection{Examples for the Scale}
\label{examples-scale}
During alignment, annotators discussed each exchange in the three videos in detail to distinguish between different levels of the scale. Two exchanges per rating of the scale from the alignment interactions are provided as illustrative examples in the video provided in the footnote\footnote{HRI CUES exemplary exchanges: \url{https://youtu.be/VmKvGM0pyec}}. Table~\ref{tab:exchange_scores} presents five of these example exchanges for one participant (S0), along with the corresponding scores and reasoning.

In \textbf{Turn 11}, the participant sounded disappointed that the robot misunderstood the question and repeats their previous inquiry. He looks at the experimenter three times (gaze aversion) and sighs before posing their second question for clarification. The participant displays low energy throughout. These behaviors indicate signs of frustration. In \textbf{Turn 8}, the robot interrupts the participant at the beginning of the turn. In response, the participant raises their head in a slow nod, signaling annoyance, and looks at the experimenter (gaze aversion). He then repeats the sentence he was previously attempting to say in a rushed manner. The participant inhales to finish the sentence, which are signs associated with boredom. In \textbf{Turn 18}, the robot interrupts the participant at the beginning of the exchange, causing the participant to stop speaking and make an awkward expression. The robot's phrasing (\quotes{That sounds X. I love Y too}) is repeated for the fifth time in the conversation, with \quotes{I love nature too} repeated for the third time. The participant nods, smiles, and asks a follow-up question in a polite manner to move the conversation forward. In \textbf{Turn 23}, the interaction proceeds smoothly with no interruptions or communication failures. The participant appears interested in the conversation, as the robot correctly interprets and responds to their question. The participant nods in affirmation and responds with high energy. In \textbf{Turn 22}, the robot interrupts the participant mid-speech. Despite the interruption, the participant exhibits rapid turn-taking in response, demonstrating immersion in the conversation. He smiles, swing sideways in their chair, and lean forward while asking the next question, which are signs that reflect excitement.

While these examples are intended to illustrate what may constitute a given level on the scale according to the signs of enjoyment relevant to the use case in this work (\tbl~\ref{tab:signs}), the interpretation can depend on the conversation context and history (\eg prior robot failures may cause user frustration), vary between participants (\eg extroverts vs. introverts, older adult vs. young adult), rely on the dominant signals in the exchange (\eg mixed signals with differing intensities), and differ across application domains (\eg games may elicit deeper immersion than casual interactions with a companion robot). As highlighted in the previous section, it is important for annotators to establish the \quoquo{baseline} behavior from which an individual may deviate during the interaction, while considering what has happened within the conversation history. \sect~\ref{sec:otherdomains} provides guidelines for adapting the scale to other domains through expert review of the relevant signals of enjoyment for each level of the scale.

\subsection{Annotation using HRI CUES}\label{annotationEvaluation}

After establishing HRI CUES, all annotators independently rated the remaining 25 videos described in \sect~\ref{data:evaluation}. The same methodology was employed as in \sect~\ref{familiarization}, with the only difference being the enjoyment scale, as HRI CUES was used instead of the initial scale. That is, the annotators rated 590 turns (174 minutes) using ELAN with both side and frontal view of the interaction (\fig~\ref{fig:rating}), viewing each exchange only once, in addition to providing an overall enjoyment score per interaction. They also provided an explanation for their overall ratings and the challenges they faced during the annotation, as in \sect~\ref{familiarization}. The annotation was conducted over 8 days. The results are reported in the next section.

\section{Results}\label{results}

Following the discussions involved in the annotator alignment that redefined the user enjoyment scale and methodology, annotators rated the remaining evaluation data (25 videos) individually. 

\subsection{Distribution of User Enjoyment}

\begin{figure}[t]
    \centering
    \includegraphics[width=0.78\columnwidth]{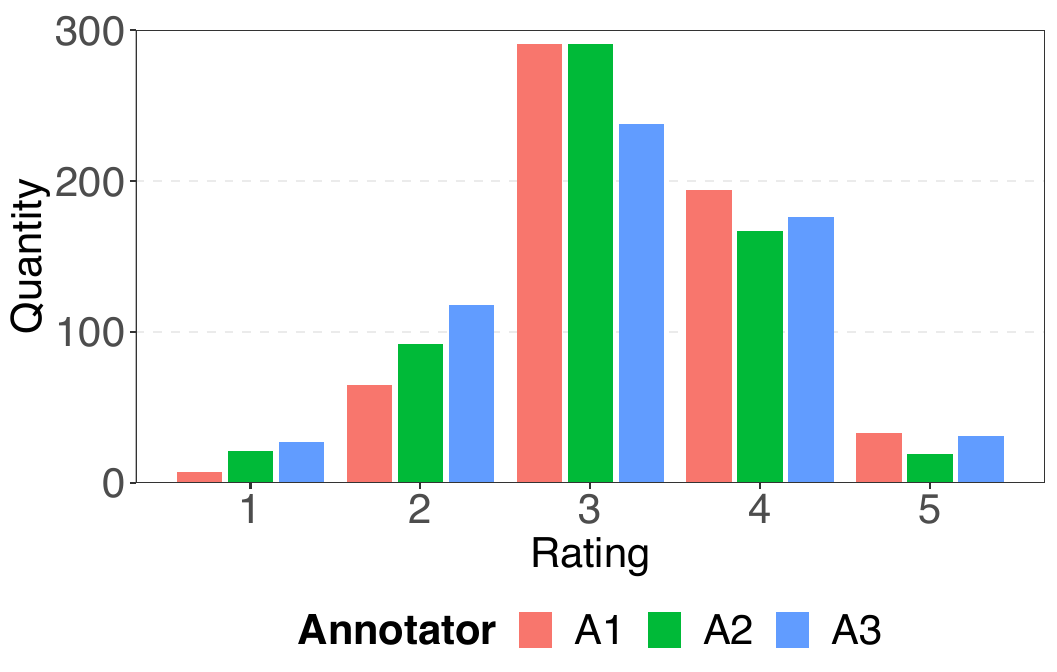}
    \caption{Distribution of HRI CUES ratings across turns of all participants per annotator.}
    \label{fig:ratingQuantity}
\end{figure}

\fig~\ref{fig:ratingQuantity} shows how each annotator rated the interaction exchanges (robot-participant turn pairs), indicating that the interactions mainly were ($45.9$\%) regarded as neutral in enjoyment, with rare occurrences of very low ($9.2$\%) and very high ($13.9$\%) enjoyment, showing a near Gaussian distribution of user enjoyment for each annotator. 
\fig~\ref{fig:ratingDistributionParticipant} (in the Appendix) shows the rating distributions of annotators per participant, which display a similar Gaussian distribution, with some participants (\eg P25, P27) perceived to have a more enjoyable interaction than others (\eg P3). 

\subsection{Rater Reliability}\label{results:reliability}
To evaluate the reliability of the annotators' enjoyment ratings, we employed the Intraclass Correlation Coefficient (ICC), similar to~\cite{reimnitz2022enjoyment}, which is a statistical measure used to assess the reliability or consistency of ratings provided by multiple raters (or annotators)~\cite{koo2016guideline}. ICC values range from 0 to 1, with higher values indicating greater agreement among raters\footnote{ICC value less than 0.5 is poor reliability, between 0.5 and 0.75 is moderate, between 0.75 ad 0.9 is good, and above 0.9 is excellent reliability~\cite{koo2016guideline}.}. This study focuses on two specific forms of ICC: 
\begin{itemize}
    \item ICC(2) - Single Random Raters: Designed for situations where each subject is rated by the same raters, and those raters are considered to be randomly selected from a larger population of possible raters.
    \item ICC(2,k) - Average Random Raters: An extension of ICC(2), applied when the average ratings of k raters are considered, enhancing the reliability of the measurement.
\end{itemize}

Similarly to how our data was coded, we present rater reliability for each conversation exchange and the overall enjoyment score provided by each annotator for the interactions. Detailed analysis of concordance and divergence between annotators is provided in the Appendix.

\subsubsection{Per Conversation Turn}
The resulting annotations per conversation turn of 25 videos are shown in \fig~\ref{fig:annotationTurns} in the Appendix. Treating each exchange as a repeated measures factor in the reliability analysis: 
\begin{itemize}
    \item The ICC(2) was $0.47$ with 95\% confidence interval ranging from $0.23$ to $0.69$, indicating poor to moderate level of reliability. This was statistically significant ($p < 0.001$) with an $F$-statistic of $3.83$ ($df_1 = 24$, $df_2 = 48$). 
    \item For the average ratings of all coders, the ICC(2,3) was $0.72$ with 95\% confidence interval of $0.47$ to $0.87$, suggesting a poor to good level of reliability, which was statistically significant ($p < 0.0001$). This was further supported by the same $F$-statistic.
\end{itemize}

Based on the visual inspection of annotator ratings (\fig~\ref{fig:annotationTurns}), A1 was identified to diverge from A2 and A3. A1 was substantially more positive ($M=3.31$ for turns) from the other annotators ($A2: M=3.12$, $A3: M=3.11$). To confirm this, we evaluated ICC with A1 excluded:
\begin{itemize}
    \item ICC(2) for single random raters rose substantially to $0.74$, with 95\% confidence interval ranging from $0.49$ to $0.88$, indicating a much stronger reliability between A2 and A3. This result was statistically significant ($p < 0.001$) with an $F$-statistic of $6.52$ ($df_1 = 24$, $df_2 = 24$). 
    \item When considering the average ratings of the remaining two annotators (k=2 in ICC), ICC(2,2) was an impressive $0.85$, with 95\% confidence interval of $0.66$ to $0.93$, suggesting moderate to excellent reliability, which was statistically significant ($p < 0.0001$). This was further confirmed by the same $F$-statistic.
\end{itemize}

These results confirmed our initial conclusion. In addition, removing A2 or A3 separately decreased ICC. Subsequent discussions with the annotators further confirmed a divergence in the ratings provided by A1 (\sect~\ref{sec:divergence}).

\subsubsection{Overall Enjoyment Score}
The overall enjoyment scores per annotator are presented in \fig~\ref{fig:userPerceptionComparisons} in the Appendix. Reliability among overall enjoyment scores was:
\begin{itemize}
    \item The ICC(2) for single random raters was found to be $0.48$, with 95\% confidence interval ranging from $0.24$ to $0.69$, indicating poor to moderate level of reliability. This value was statistically significant ($p < 0.001$) with an $F$-statistic of $3.74$ ($df_1 = 24$, $df_2 = 48$). 
    \item The average ratings from three annotators (ICC(2,3)) was $0.73$, with 95\% confidence interval of $0.48$ to $0.87$, suggesting a poor to good level of reliability, which was statistically significant ($p < 0.0001$) with $F$-statistic of $3.74$ ($df_1 = 24$, $df_2 = 48$). 
\end{itemize}

Similar to per-turn analysis, excluding the divergent annotator (A1) led to improved reliability: 
\begin{itemize}
    \item ICC(2) increased to $0.58$, with 95\% confidence interval ranging from $0.25$ to $0.79$, suggesting a higher consistency between two annotators. This was statistically significant ($p < 0.001$) with an $F$-statistic of $3.74$ ($df_1 = 24$, $df_2 = 24$).
    \item When considering the average ratings of the remaining two annotators, the ICC(2,2) for average random raters was $0.74$, with 95\% confidence interval of $0.4$ to $0.88$, further indicating enhanced reliability. This was statistically significant ($p < 0.001$) by the same $F$-statistic of $3.74$ ($df_1 = 24$, $df_2 = 24$).
\end{itemize}

These findings underscore the importance of selecting consistent raters and the benefit of averaging ratings across multiple annotators to achieve enhanced reliability in measuring user enjoyment in robot conversations.

\subsection{Annotator Correlations with Self-Reported Enjoyment}
Users' subjective ratings of enjoyment during the interactions were obtained from the questionnaire in the second study after their interaction with the robot, in terms of user satisfaction, fun, interestingness, and strangeness of the conversation, as described in \sect~\ref{data:secondStudy}, which are presented in \fig~\ref{fig:userPerceptionComparisons} (in the Appendix) along with the annotator overall enjoyment ratings per interaction. The items had high reliability (Cronbach's alpha = $0.84$), with the removal of each item reducing the correlation in the construct. These self-reported scores and the average of these scores were compared against annotators' overall enjoyment scores to evaluate how well the annotators could perceive their enjoyment. Spearman correlation was used across four Likert scale items and the average of these scores, with 95\% confidence interval (ranging from $0.71$ to $0.92$). 
The results were as follows:

\begin{itemize}
\item \textbf{Overall vs. User Average:} Not statistically significant ($p = 0.08$) moderate positive correlation ($r = 0.36$),
\item \textbf{Overall vs. User Satisfaction:} Not statistically significant ($p = 0.06$) moderate positive correlation ($r = 0.39$), 
\item \textbf{Overall vs. User Fun Talking:} Not statistically significant ($p = 0.21$) weak positive correlation ($r = 0.26$),
\item \textbf{Overall vs. User Conversation Interesting:} Not statistically significant ($p = 0.68$) very weak positive correlation ($r = 0.09$),
\item \textbf{Overall vs. User Felt Strange (Reversed):} \textit{Statistically significant} ($p = 0.04$) moderate positive correlation ($r = 0.42$).
\end{itemize}

Detailed analysis of similarity and discrepancy between annotators and user perceptions is provided in the Appendix \sect~\ref{sec:discrepancies}.
 
\subsection{Large Language Model Correlations with Self-Reported Enjoyment}
HRI CUES was further examined for its potential in automatic enjoyment detection. To this end, the turn-by-turn enjoyment ratings generated by various LLM configurations (for more details see~\cite{pereira2024multimodal}) using the scale were averaged for each interaction and correlated with the same participants' post-interaction self-reported ratings.

The results show a more notable alignment between LLM assessments and user perceptions. The GPT-4 Turbo model configured with all prompt features (including scale, instructions, examples, conversation history, and Chain-of-Thought reasoning) achieved the following statistically significant correlations with user self-reported ratings: 
\begin{itemize}
\item \textbf{LLM Average vs. User Average:} \textit{Statistically significant} ($p = 0.002$) and strong positive correlation ($r = 0.60$).
\item \textbf{LLM Average vs. User Satisfaction:} \textit{Statistically significant} ($p < 0.001$) and strong positive correlation ($r = 0.68$).
\item \textbf{LLM Average vs. User Fun Talking:} \textit{Statistically significant} ($p = 0.026$) and moderate positive correlation ($r = 0.44$).
\item \textbf{LLM Average vs. User Conversation Interesting:} Not statistically significant ($p = 0.073$) moderate positive correlation ($r = 0.37$).
\item \textbf{LLM Average vs. User Felt Strange (Reversed):} \textit{Statistically significant} ($p = 0.017$) and moderate positive correlation ($r = 0.47$) .
\end{itemize}
Other LLM configurations, including multimodal models like Gemini Pro 1.5, also showed significant moderate correlations, particularly with satisfaction and strangeness. 

LLMs have also shown the ability to predict turn-by-turn user enjoyment with HRI CUES even without access to the user's response to a robot's utterance~\cite{janssens2025online}. The turn-by-turn exchange level predictive capability performs comparably to enjoyment detection (with user response) from the human experts annotations reported here. On a correlation level, even when removing user responses, GPT‑4o still achieved significant correlation values of ($r = 0.59, p = 0.01$) with satisfaction and ($r = 0.54, p = 0.03$) with strangeness (reversed).






\section{Discussion}

\subsection{Enhanced Reliability with Averaged Annotator Ratings}
While having a similar distribution of ratings by all annotators, the reliability analysis revealed a marked distinction between ICC(2), which assesses the reliability of single random raters, with ICC(2,k), which considers the average ratings of multiple annotators. The latter consistently demonstrated higher reliability across both overall enjoyment scores and conversational turn ratings. This finding highlights the substantial benefit of collective annotator wisdom over individual assessments in assessing user enjoyment. 

The superior reliability of ICC(2,k) highlights the inherent variability in subjective experiences and perceptions of enjoyment, suggesting that averaging across multiple annotators can effectively mitigate individual biases and variations in judgment. Divergence of annotator perceptions based on the context and cues of the exchange (detailed in \sect~\ref{sec:divergence} in Appendix) aligns with challenges detected for user satisfaction annotations~\cite{Higashinaka2010IssuesIP}, suggesting that averaging across annotators can lead to a more reliable representation of true user enjoyment. This finding is critical, as it emphasizes the importance of incorporating multiple perspectives to achieve a more accurate and consistent evaluation of user enjoyment in conversational interactions with robots. Consequently, the distinction between ICC(2) and ICC(2,k) results not only supports the robustness of our user enjoyment scale but also illustrates the methodological importance of employing multiple annotators for capturing the complex and subjective nature of enjoyment in human-robot conversations.

\subsection{Correlation with User Perceptions}
To further evaluate our user enjoyment scale, we analyzed correlations between the annotators' overall enjoyment scores and the participants' subjective enjoyment ratings. Results revealed a statistically significant, moderate positive correlation between overall enjoyment scores and the (reversed) \quoquo{felt strange} item, indicating that higher enjoyment scores were associated with decreased feelings of awkwardness during the interaction. This appears to greatly align with our annotator's discussions and the resulting enjoyment scale that classifies the presence of signs of discomfort as the scale's lowest level.

However, the correlations between annotator scores and other user-reported measures — such as satisfaction, average enjoyment, fun in talking, and interest in conversation — although trending towards moderate positive correlations, did not reach statistical significance. These findings suggest a nuanced relationship between observed and user experiences (detailed in \sect~\ref{sec:discrepancies} in Appendix). While annotators can detect general levels of comfort and ease within interactions, capturing comprehensive internal subjective enjoyment may require additional data, such as initial expectations towards the robot, personality traits, as well as physiological responses, not accessible through direct observation. This discrepancy underscores the complexity of correlating observed behavior with subjective internal states and highlights the challenge of fully capturing user enjoyment in HRI.

However, complementary evidence comes from our LLM analysis. GPT-4 Turbo, guided by the full HRI CUES  prompt, achieved strong and statistically significant correlations with most self-ratings, including for the user-average score, and even the GPT-4o variant—without access to the user’s replies—retained significant correlations. Multimodal models such as Gemini Pro 1.5 showed comparable, moderately strong alignments. These results indicate that evaluating enjoyment with HRI CUES using automated systems, such as foundation models, can reveal subjective enjoyment cues that human observers may overlook, likely due to the models' ability to integrate subtle lexical, prosodic, and contextual patterns throughout the dialogue. This finding aligns with prior literature on automatic assessment outperforming annotators in correlations with user perceptions~\cite{Wei2021multimodal}.  The convergence between self-ratings and LLM predictions points to a path toward real-time, automated enjoyment monitoring during open-domain dialogue in HRI. 

Our approach aligns with how humans naturally adapt our conversations in real-time based on the external cues we observe in others. The measure presented in this paper offers a valuable tool for collecting observational data to train autonomous enjoyment detection systems that can be used on robots or other agents.
 
\subsection{Adapting HRI CUES to Other Domains}\label{sec:otherdomains}
A final round of annotator discussions (lasting 3 hours) was conducted to adapt the developed scale to other domains in HRI, extending its application beyond the context of companion robots for older adults to encompass other conversational contexts where analyzing enjoyment is crucial.

The process of using the Human-Robot Interaction Conversational User Enjoyment Scale (HRI CUES) is twofold. First, the scale is the primary supportive instrument for engaging with the dataset and finding an agreement among annotators regarding which level of enjoyment an exchange represents. Secondly, it is important to highlight the cultural and context-dependent changes when assessing enjoyment with the scale, therefore, we recommend that the annotators reach an agreement on which multimodal cues are important in their study. In the previous section, we presented the relevant cues in our study located in a Nordic Western setting, but these could differ in another setting. For example, gestures, such as thumbs up, mean different things in different parts of the world. Our proposed scale can already be directly applied in many settings where users engage with social robots (or potentially other agents) in conversations. However, in many other cases, we encourage further adapting the scale to the particular domain.

To replicate our methodology and adapt HRI CUES to another domain (assuming that a clear understanding of the intended users, domain, and context is already established), the following framework should be applied:

\begin{enumerate}
    \item Recruit three annotators with relevant and complementary backgrounds who are familiar with the specific culture and context of the study.
    \item Establish the intended usage of the annotations such that the annotators can tailor their annotations to fit that use case and align their views on the practical meaning of enjoyment in that context.
    \item Ask annotators to systematically annotate three example videos (from the dataset) using the HRI CUES. Example videos should exemplify different interaction outcomes from the dataset and be represented from various angles (front, side, both). Encourage the annotators to look for their own cues of enjoyment or dis-enjoyment, and describe the reasons behind their overall enjoyment scores based on those after viewing each interaction, and the corresponding challenges of detecting enjoyment. 
    \item Arrange a discussion between the annotators to identify contrasting and supportive arguments for multimodal cues associated with the scale, aiming to precisely determine the cues for each segment and strive for consensus for the corresponding rating. Give each segment sufficient time for discussions while avoiding getting stuck on small details. When faced with a difficult case, note what is not agreed on and move to the next segment.
    \item Based on the discussion, construct an annotation schema, which should contain the cues that were agreed on for assessing the enjoyment in relation to the scale, especially emphasizing the cues that were discussed and not immediately agreed on.
    \item Annotate the remaining dataset turn-by-turn using the HRI CUES and the multimodal cues. This is done by looking at each exchange once and annotating in real-time without going back in the data to not influence the evaluation of the beginning segments of the interaction by already knowing the end segments.
\end{enumerate}
In this paper, we evaluated HRI CUES through the interactions of older adults with a conversational companion robot using Furhat at the university premises. 
However, this was only one example of how the scale can be used; HRI CUES is generalizable to other contexts in which a user interacts with a social robot. Therefore, the second to fifth steps of the framework above are crucial for the annotators to adapt the communication cues to their context and setting of the study. HRI CUES does not require any adjustment as such, but it requires interpretation with respect to the context of the use case. The interpretation is facilitated by the six-step framework, hence, it is important to find appropriate annotators who are familiar with the particular application area.

\subsection{Challenges and Limitations}\label{sec:challenges}

In this work, we introduced a novel scale for annotators to evaluate user enjoyment in conversational HRI. However, enjoyment is a subjective measure, and thus, is challenging to evaluate and agree on between annotators and correlate with users perceptions, given the multitude of aspects connected to it. For instance, the context and length of the exchange affected how the enjoyment was perceived. 
When segments were overly brief, the exchange did not always contain sufficient information for a fair assessment. More frequently, however, excessively long segments were complex to analyze. For example, the annotators often interpreted the exchange differently due to long segments that contained several cues belonging to separate levels of the scale (described in \sect~\ref{sec:divergence}). In these cases, a different assessment approach may be necessary as the longer segments introduced an additional factor for the annotators to consider: which part or aspect of the exchange to emphasize in the assessment. Due to the complex nature of enjoyment as a concept, which highlights the necessity of pre-coding discussions among the annotators, aligning on the definition of enjoyment of relevance to the particular study is challenging. One solution is to focus on the change of behavior within the exchange. For instance, if the robot's response improved the user's demeanor towards the robot, the exchange should be rated towards the level that contains higher enjoyment cues in the scale, and conversely if it had a negative impact. Other alternatives could be to use an aggregated score, the rating that corresponds to the majority of the segment, the most/least extreme rating, or the rating that corresponds to the first or last part of the segment. The choice of method for these situations should be determined during the annotation alignment process, taking into account the specific use case and domain.

The differences between the annotators' overall rating on user enjoyment and user perceptions (detailed in \sect~\ref{sec:discrepancies} in Appendix) might be due to the participant's expectation of the robot's social and technical level, while the annotators only look at the interaction itself. In addition, since this is their first interaction with a robot for most of the participants (20 out of 25), the \quoquo{novelty effect} might have changed their perceptions more positively or negatively, given that the duration (7 minutes) is not long enough to overcome it~\cite{jost2020hri}. 
However, the variability observed by the annotators in user enjoyment states throughout the interaction (\eg \fig~\ref{fig:divergence} in Appendix) shows that conversation context may alter the novelty effect, providing a more complete picture of the enjoyment throughout the interaction than a self-reported score at the end of the interaction. In addition, the users' responses to the questionnaire may differ from their actual attitudes towards the robot~\cite{reimann2024social}. These support the importance of HRI CUES as an additional tool to evaluate user enjoyment, providing means for real-time estimation of enjoyment in conversational agents.

A limitation worth noting was that the scale was developed and evaluated on open-domain conversations with a companion robot with older adults from Sweden. However, the scale development was initially based on enjoyment in marriage conversations between couples~\cite{reimnitz2022enjoyment}, in addition with various multimodal cues derived from the HRI and HCI literature in affective computing, user engagement, user enjoyment, conversation, and turn-taking analysis. In other words, HRI CUES levels or signs of enjoyment were not specific to older adults or the context of the interaction. Thus, the scale is generalizable to other contexts in which a user interacts with a social robot. 
In addition, we offered guidelines to adapt the scale to other application domains with robots or agents (\sect~\ref{sec:otherdomains}). Nonetheless, future work is necessary to validate the scale in other domains~\cite{li2010cross}, with a larger sample~\cite{bethel2009use}, participants from other cultures and socio-economic backgrounds~\cite{haring2014cultural, li2010cross}, a broader age range~\cite{chien2019age} and varying levels of familiarity with technology and robots~\cite{bishop2019social}, and in long-term deployments to mitigate the novelty effect~\cite{degraaf2016long}, as users' expectations and enjoyment of robot interactions may vary depending on individual differences, contextual factors, and the duration of interaction. 

Our scale is designed to serve as a tool for assessing perceived user enjoyment during interactions with robots, intended for researchers in the field of HRI and automatic systems for real-time or offline evaluation. However, given that open-domain dialogue may involve sensitive information that individuals may be hesitant to share with unfamiliar parties or automated systems, it is imperative to obtain explicit consent from participants before they talk with the robot. In our studies, we ensured this consent prior to participants' interactions with the robot, employing the use of the term \quoquo{sentiment analysis} in the consent form, and explained to all participants that their interactions would be analyzed in terms of their affective states (\quoquo{identification of feelings}) by both researchers and automated systems. While obtaining consent is essential for researchers utilizing our scale in future studies, it is crucial to recognize that this process may influence participants' behavior and conversation topics, as they may be reluctant to share sensitive memories or be concerned about being judged by others. Consequently, this can result in a disconnection with the robot, posing challenges in achieving high levels of enjoyment during interactions. Nonetheless, this impact is likely to diminish over the course of the interaction or across multiple interactions, particularly in long-term settings. Additionally, researchers must exercise caution to uphold the privacy and confidentiality of participants when sharing data, ensuring that they remain unidentifiable in images and videos (as demonstrated in the examples provided for the scale), and removing any sensitive information. Moreover, researchers should remain mindful of their own biases and subjectivity, which may lead to variations in the interpretation of enjoyment compared to the participants' experiences. User enjoyment is often context-specific, indicating that users' behavioral and affective expressions are connected to specific socio-cultural contexts, including values, norms, and expectations of what is considered appropriate in certain situations~\cite{van2016social}. These underscore the significance of employing multiple annotators for the scale that are familiar with the socio-cultural background of the target population, and using the scale to complement self-reported user perceptions to have a deeper understanding of the interaction quality. As human-robot interactions continue to evolve, ensuring a deep understanding of user enjoyment not only elevates the quality of these interactions but also paves the way for more empathetic and meaningful connections. 

\section{Conclusion}
Our research contributed a novel scale for measuring user enjoyment in conversations with a robot from an external perspective. The scale was developed through rigorous discussions of three annotators with complementary and relevant backgrounds to user enjoyment and the application domain. Older adults' interactions with a companion robot were used as the basis for developing the scale, which was evaluated on 174 minutes of interactions of 25 participants. Inter-rater reliability analysis showed the importance of using multiple annotators, with moderate to good alignment, where the disagreements arose from the complexity and the subjectivity of user enjoyment, especially when a user shows various signs of enjoyment and dis-enjoyment within a single conversation exchange. The overall user enjoyment rated per interaction correlated significantly with users' perceived level of strangeness of the conversation, which signifies that the (dis)comfort experienced in the interaction was correctly identified by the annotators, and shows the importance of including dis-enjoyment levels in the scale. Using our scale with LLMs has been shown to correlate even more strongly with users' self-reported ratings, paving a promising path toward automation.
These findings support that our user enjoyment scale is a viable measure for external assessment of user enjoyment, and emphasize the critical role of methodological rigor in assessing subjective experiences within conversational robot interactions. Our study emphasizes the value of using multiple annotators and proposes potential scale refinements to further enhance consistency in quantifying the nuanced concept of enjoyment across application domains. The developed scale and the corresponding dataset aim to provide a tool for measuring user enjoyment from an external perspective to supplement self-reported user enjoyment responses in HRI research, with future potential application for autonomous detection of user enjoyment in real-time in robots and agents for adapting conversations contingently to provide enjoyable and long-lasting interactions.


\section*{Acknowledgments}
We would like to thank Aida Hosseini for manually correcting transcripts of robot interactions, and 
the study participants  
for their time and efforts.

\bibliographystyle{IEEEtran}
\bibliography{sample-base}

\newpage

\section{Biography Section}

\begin{IEEEbiography}[{\includegraphics[width=1in,height=1.25in,clip,keepaspectratio]{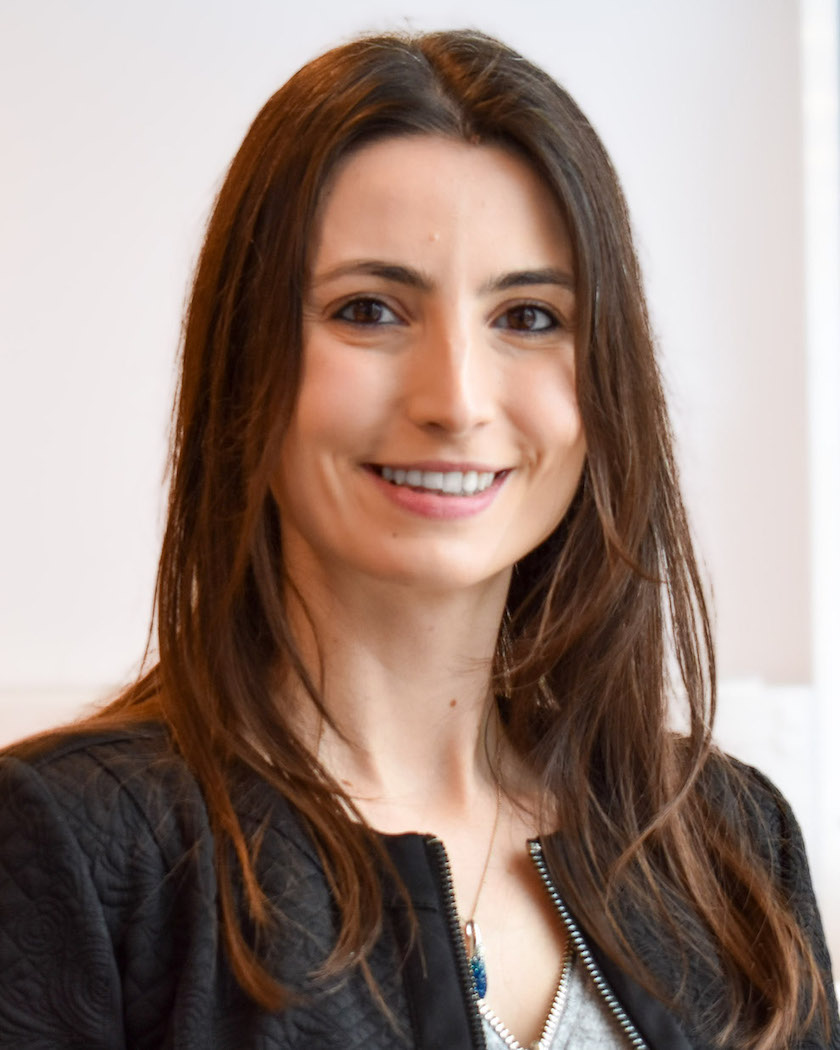}}]{Bahar Irfan}
is a Postdoctoral Researcher and Digital Futures fellow at KTH Royal Institute of Technology, Sweden. Her research focuses on creating personal robots that continually learn and adapt to assist in daily life. Her research interests include social robots, lifelong learning, and large language models. Prior to joining KTH, she held research positions at Evinoks and Disney Research. She received her PhD (2020) in robotics from the University of Plymouth and SoftBank Robotics Europe as a Marie Skłodowska-Curie Actions fellow. She has an MSc (2016) in Computer Engineering and a BSc (2012) in Mechanical Engineering from Bo\u{g}azi\c{c}i University. 
Contact her at birfan@kth.se. Website: https://baharirfan.com
\end{IEEEbiography}

\vspace{11pt}

 \begin{IEEEbiography}[{\includegraphics[width=1in,height=1.25in,clip,keepaspectratio]{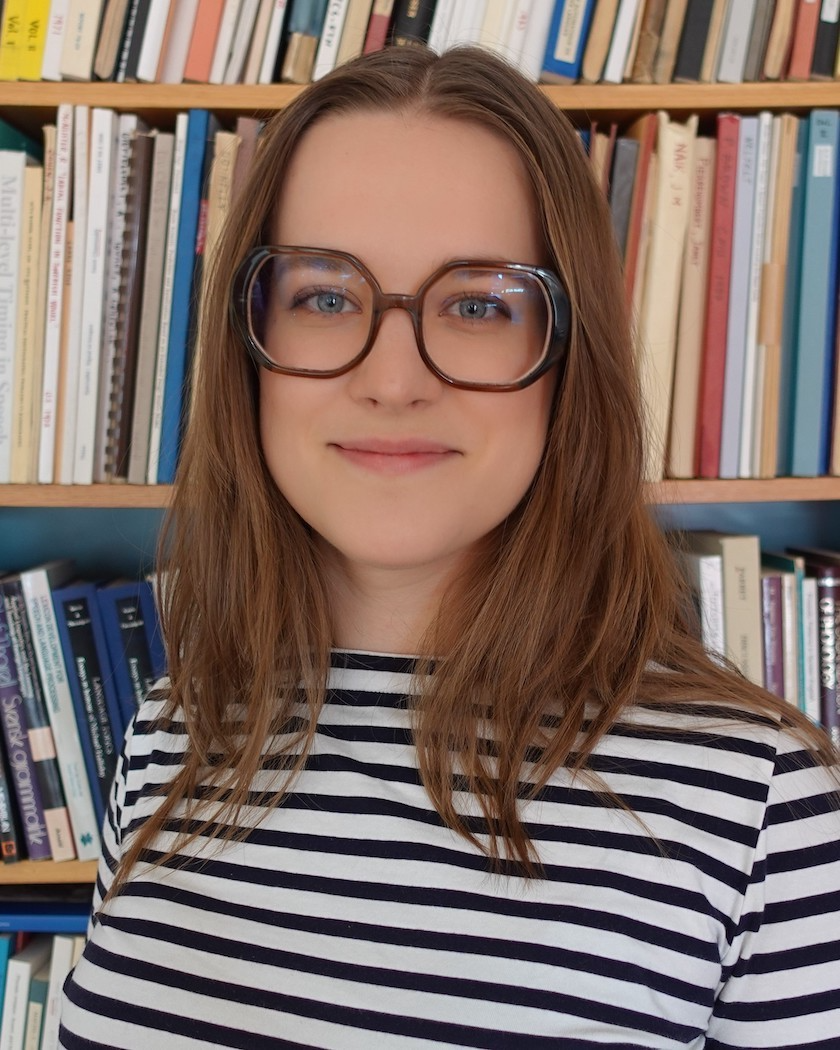}}]{Jura Miniota}
is a PhD candidate at KTH Royal Institute of Technology in Stockholm, Sweden. In her research, she is focusing on the development of socially embodied and socially intelligent robots that are enjoyable to interact with. Her research interests include multimodal artificial intelligence, social robotics, and designing enjoyable interactions. Miniota received her M.Sc. in Interactive Media Technology from KTH. Contact her at jura@kth.se. Website: https://www.juraminiota.com
 \end{IEEEbiography}

\vspace{11pt}

 \begin{IEEEbiography}[{\includegraphics[width=1in,height=1.25in,clip,keepaspectratio]{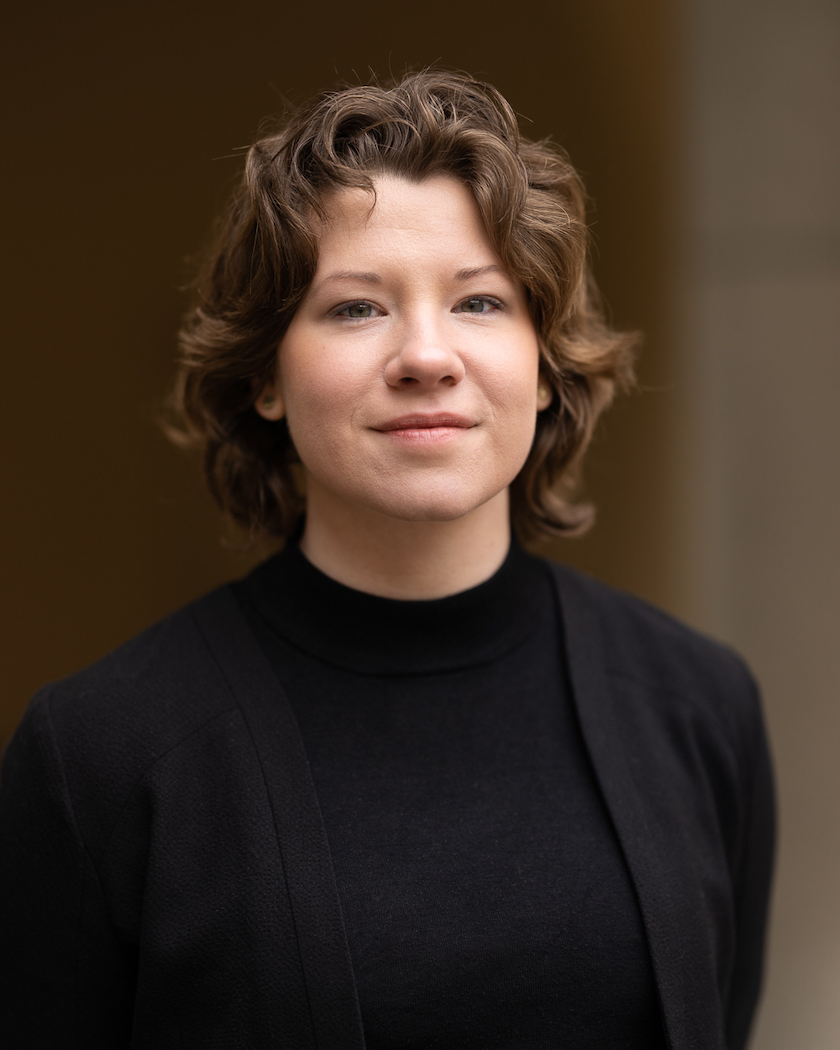}}]{Sofia Thunberg}
is Postdoctoral Researcher at the Department of Computer Science and Engineering, Chalmers University of Technology and Gothenburg University, Sweden. Her research focuses on companion robots for older adults with cognitive impairments and children with autism spectrum disorder. Her research is mainly conducted in the field, taking on a holistic approach to interactions with robots in the real world. Thunberg received her PhD in Cognitive Science from Linköping University. Contact her at sofia.thunberg@chalmers.se.
 \end{IEEEbiography}

\vspace{11pt}

\begin{IEEEbiography}[{\includegraphics[width=1in,height=1.25in,clip,keepaspectratio]{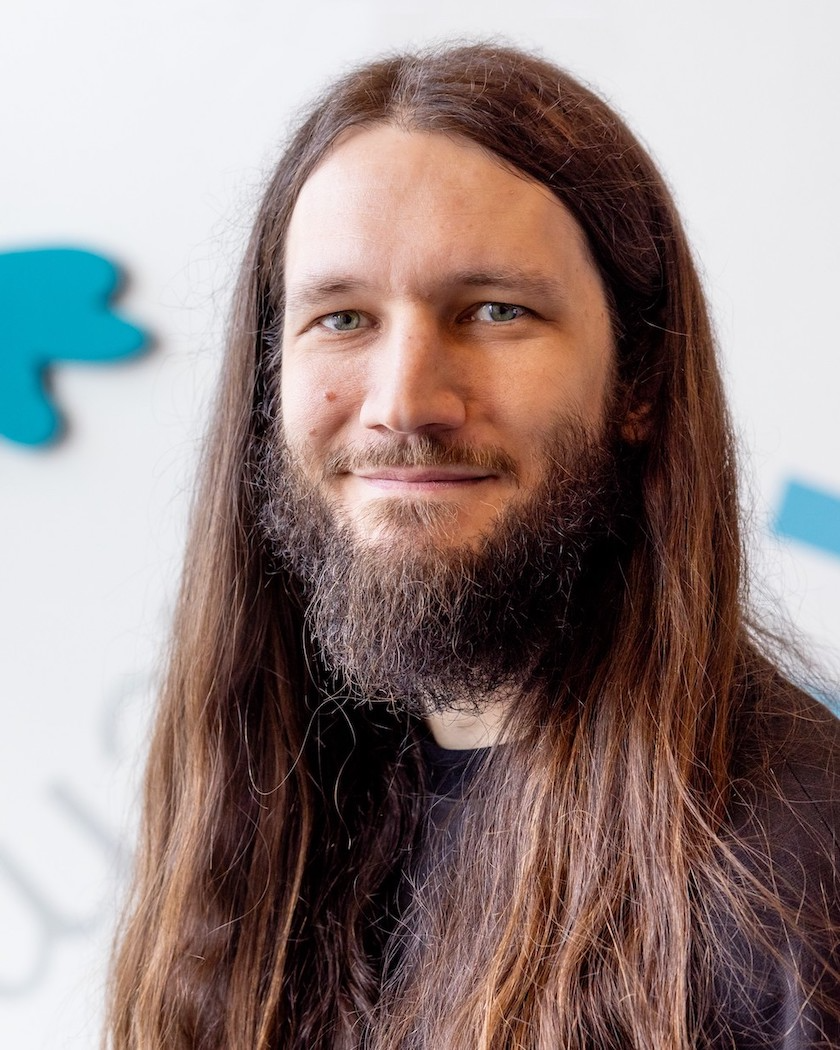}}]{Erik Lagerstedt}
is a Postdoctoral Researcher in Computational Linguistics at the University of Gothenburg, Sweden. His research interests include the evaluation and application of social robots, the theoretical foundations of interaction with technology, and the societal impact of decisions in the design and deployment of specific technologies. Lagerstedt received his PhD (2024) in Informatics from the University of Skövde, Sweden. Contact him at erik.lagerstedt@gu.se.
\end{IEEEbiography}

\vspace{11pt}

\begin{IEEEbiography}[{\includegraphics[width=1in,height=1.25in,clip,keepaspectratio]{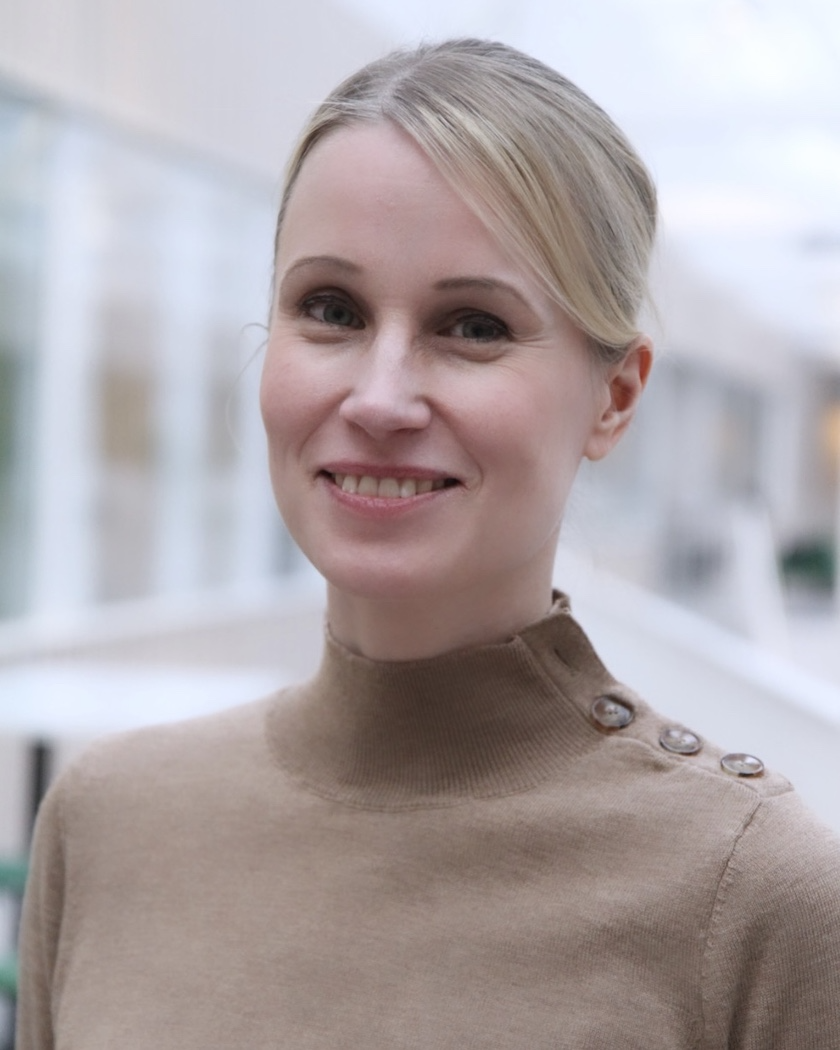}}]{Sanna Kuoppamäki}
is an Assistant Professor at the Department of Biomedical Engineering and Health Systems, KTH Royal Institute of Technology, Sweden. Her research explores the use and design of interactive technologies for healthy ageing from the age and life course perspective, with application areas in social robotics, welfare technology and mobile health. She received her PhD in sociology from the University of Jyväskylä, Finland. She is a member of the Socio-Gerontechnology network. Contact her at sannaku@kth.se
\end{IEEEbiography}

\vspace{11pt}

\begin{IEEEbiography}[{\includegraphics[width=1in,height=1.25in,clip,keepaspectratio]{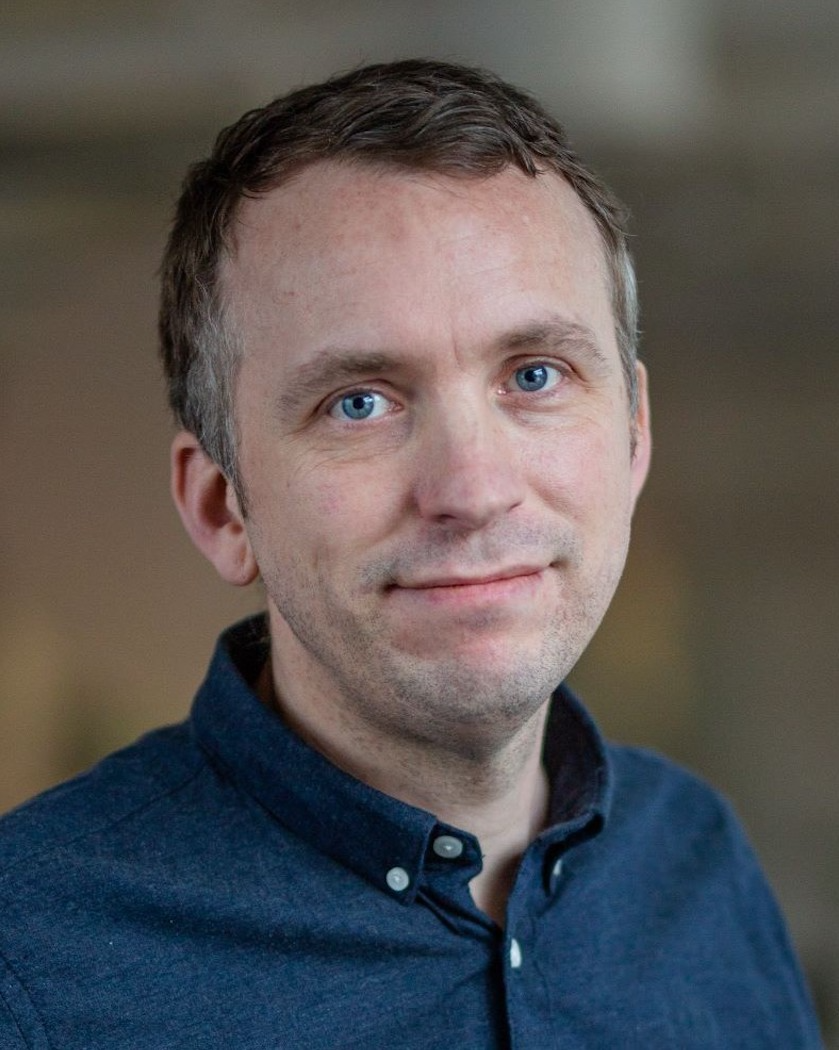}}]{Gabriel Skantze}
is a Professor in Speech Technology and Communication at KTH Royal Institute of Technology, Sweden. He is leading several research projects related to conversational systems and human-robot interaction, investigating and modeling phenomena such as turn-taking, visual grounding, and multimodal feedback in dialogue. He is also co-founder and chief scientist at Furhat Robotics and President Emeritus of SIGDIAL, the ACL Special Interest Group on Discourse and Dialogue. Contact him at skantze@kth.se
\end{IEEEbiography}

\vspace{11pt}

\begin{IEEEbiography}[{\includegraphics[width=1in,height=1.25in,clip,keepaspectratio]{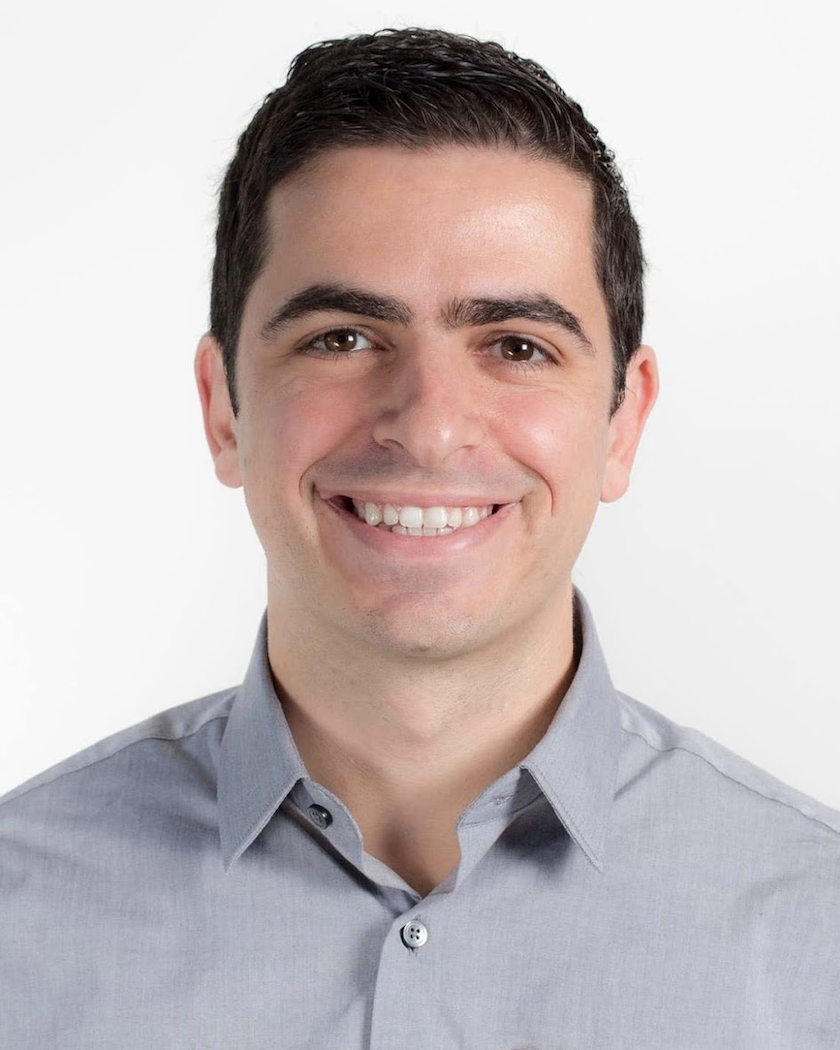}}]{André Pereira}
is a Human-Robot Interaction researcher at KTH Royal Institute of Technology, Sweden. He received his PhD in Computer Engineering from the Technical University of Lisbon, Portugal. After his PhD and a postdoctoral position at Yale University, he worked in industry at Disney Research and Furhat Robotics. At KTH, he continues to design and develop embodied socially intelligent agents for use in entertainment, education, and healthcare. Contact him at atap@kth.se
\end{IEEEbiography}

\vfill

\clearpage

\appendix

 \begin{figure*}[b]
    \centering
    \includegraphics[width=0.98\textwidth]{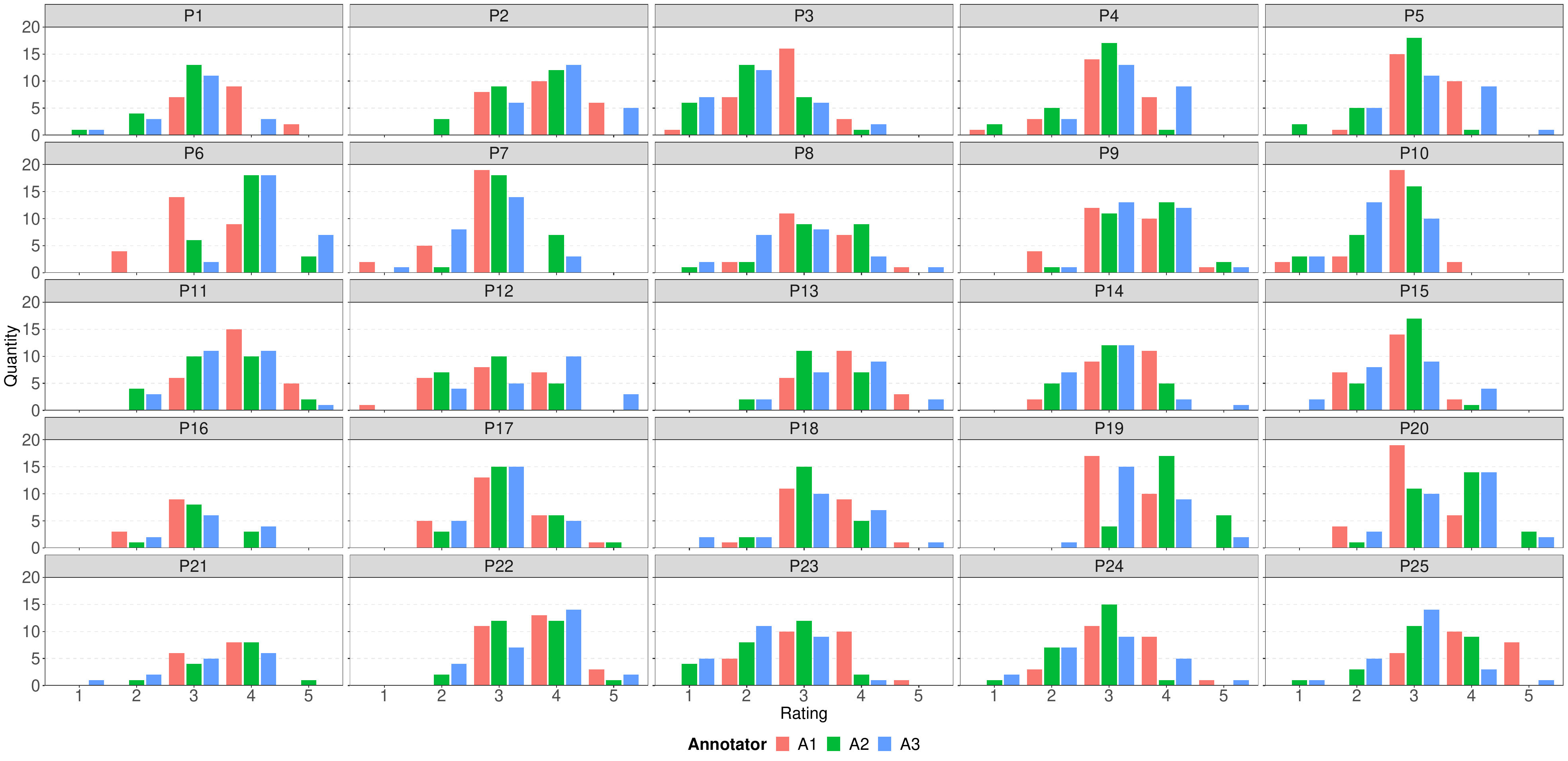}
    \caption{Rating distributions of the annotators per participant.} 
    \label{fig:ratingDistributionParticipant}
\end{figure*}

 \begin{figure*}[b]
    \centering
    \includegraphics[width=0.98\textwidth]{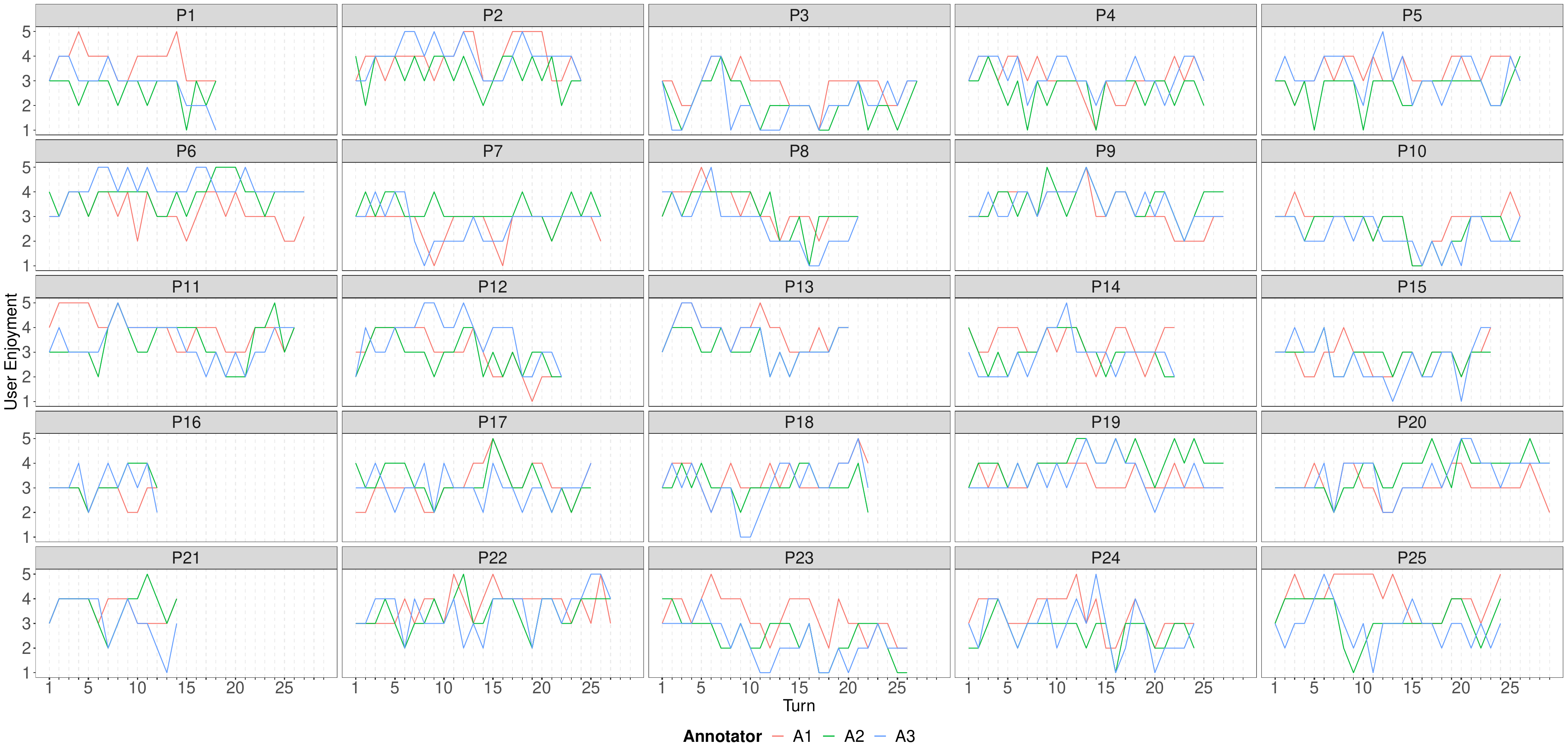}
    \caption{Annotators' user enjoyment ratings per turn for each participant's interaction.} 
    \label{fig:annotationTurns}
\end{figure*}

\section*{Analyzing Annotator and User Alignment}
\begin{figure*}[t]
    \centering
    \includegraphics[width=0.98\textwidth]{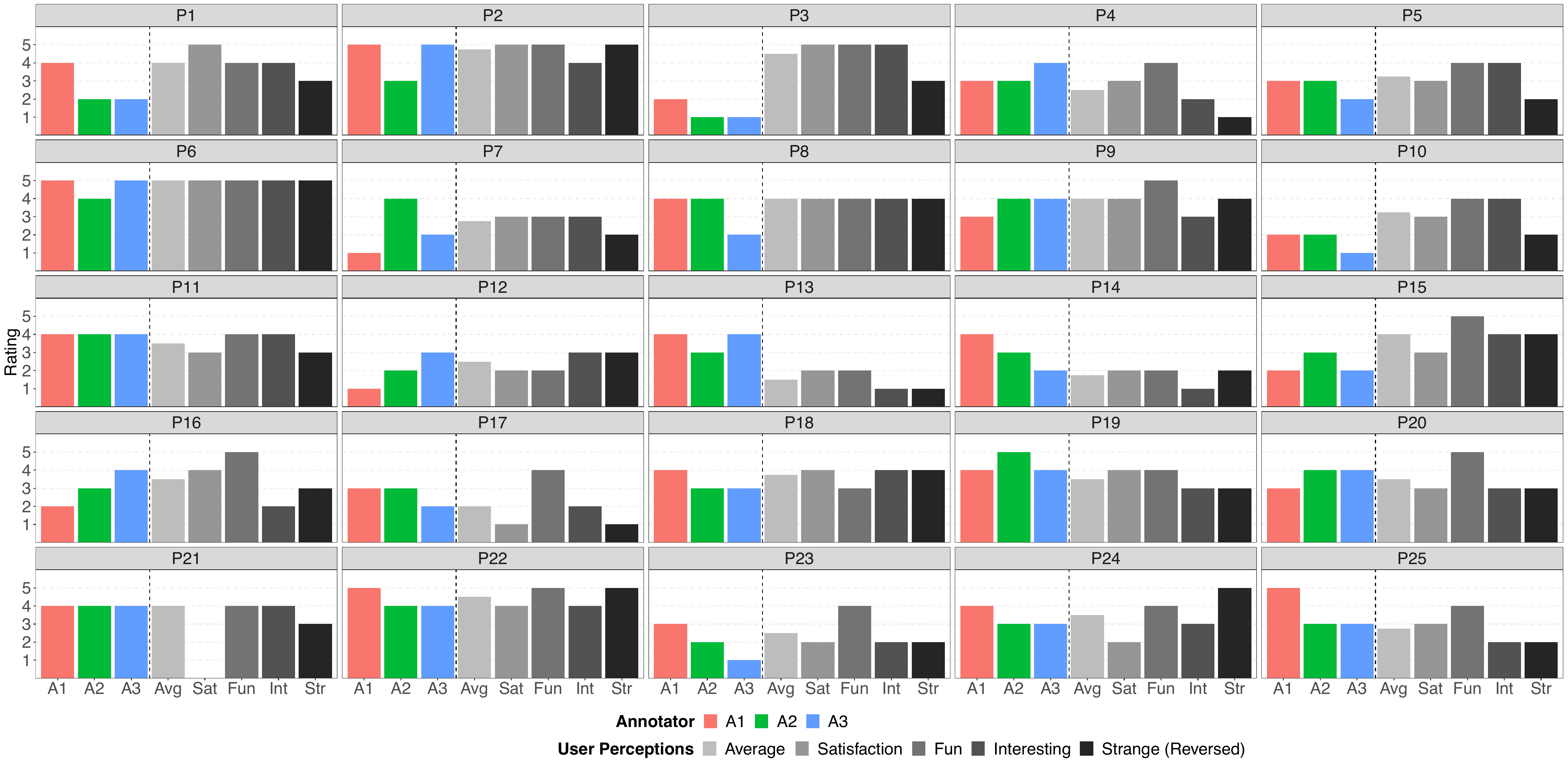}
    \caption{Annotators' ratings of the overall user enjoyment per interaction, and users' perceptions from the questionnaire.} 
    \label{fig:userPerceptionComparisons}
\end{figure*}

After analyzing the inter-rater reliability and annotator correlations with user perceptions, the annotators were asked to have another set of discussions by presenting them with \fig~\ref{fig:annotationTurns} and \ref{fig:userPerceptionComparisons} (without reliability or correlation scores), which lasted four hours. They were asked to pinpoint instances of rating divergence and concordance within their assessments. Particular emphasis was placed on identifying turns where annotators disagreed or agreed most fervently. Subsequently, the annotators were prompted to watch the corresponding video segments to explore the reasons behind their ratings and the rationale for their agreement or disagreement. The primary objective was to gain insight into potential major concordance (\sect~\ref{sec:concordance}) and divergence (\sect~\ref{sec:divergence}) in the way dialogue exchanges were annotated and elucidate the underlying reasons for these variations.

Following the turn-by-turn analysis, annotators were asked to identify the two most significant discrepancies (highest and lowest) between their ratings and self-reported participant perceptions (\sect~\ref{sec:discrepancies}). They were asked to engage in discussions exploring potential reasons underlying the disparities and similarities between their perceptions and those of the users from multiple aspects based on their expertise.

\subsection{Concordance Between Annotators}\label{sec:concordance}

\begin{figure}[t]
    \centering
    \includegraphics[width=0.93\columnwidth]{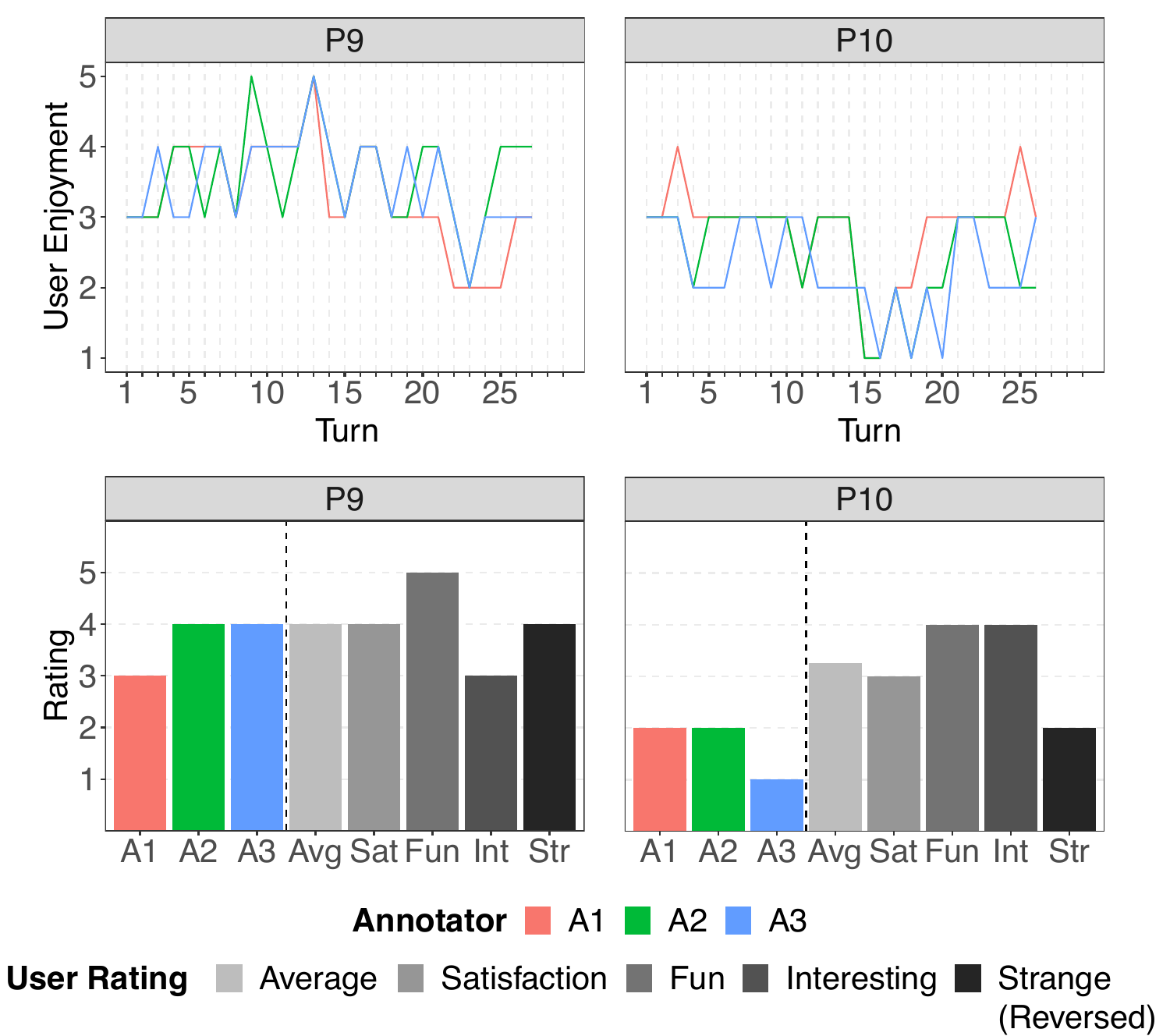}
    \caption{Concordance between annotator ratings in interactions (P9 and P10).}
    \label{fig:concordance}
\end{figure}

In numerous instances across the videos, a concordant agreement was observed among the annotators. As an illustrative example, in the interaction of the ninth participant (P9, turn 13 in \fig~\ref{fig:concordance}), the conversation exchange exhibited a seamless progression, and the participant's enjoyment level was distinctly conveyed through expansive bodily gestures. Notably, all annotators unanimously assigned a rating of five for the exchange since the participant threw themselves backward in the chair laughing. In the same video, the annotators all assigned a rating of two for the exchanges where the participant's response was marked by sighing and a demeanor suggestive of resignation. This reaction occurred as a response to an unnecessary repetition initiated by the robot, specifically at turn 23.


In another example (P10, \fig~\ref{fig:concordance}), the annotators assigned a rating of 1 to turn 16 to indicate that the participant openly expressed their negative thoughts due to the robot not making eye contact with the participant. During this interaction, the participant also attempted to establish contact with the experimenter. Subsequently, 
the participant made an effort to politely 
maintain a dialogue with the robot according to social norms, a behavior that garnered consensus among the annotators as being representative of a rating of 3.

In other words, the annotators agreed when the user enjoyment scale aligned clearly with participant behavior. However, in most cases, the interaction between the robot and the participant did not correspond as clearly or unambiguously to the user enjoyment scale. This is likely due to the complex and situational nature of the cues in the interaction, making it challenging to develop comprehensive yet precise guidelines for annotation. Instead, the general scale needs to be interpreted by the annotators for the particular use case to find anchor points that are appropriate for the specific context.

\subsection{Divergence Between Annotators}
\label{sec:divergence}

\begin{figure}[t]
    \centering
    \includegraphics[width=0.93\columnwidth]{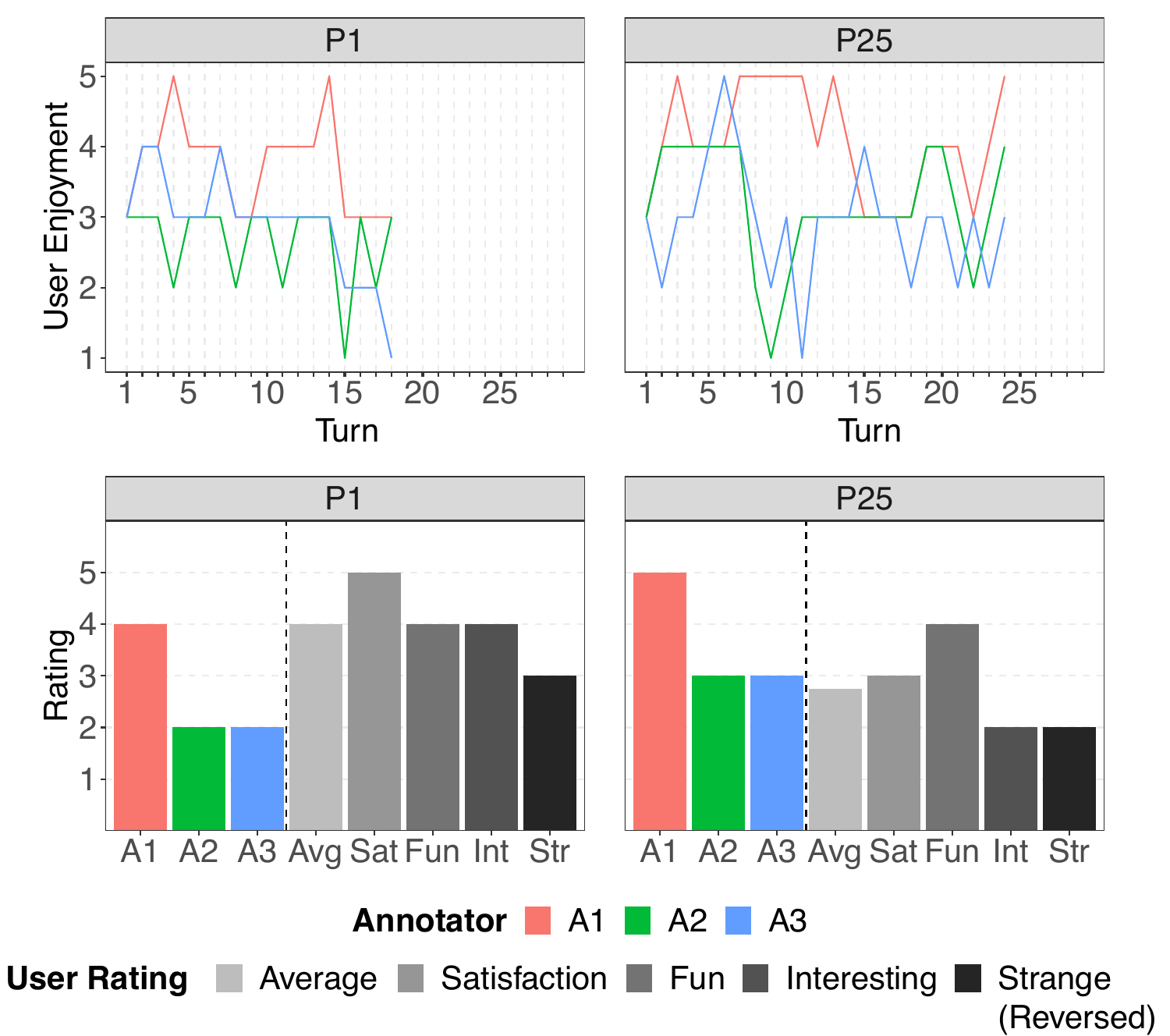}
    \caption{Divergence between annotator ratings in interactions (P1 and P25).}
    \label{fig:divergence}
\end{figure}

Throughout the analysis of the 25 videos, there were instances where annotators differed substantially in their assessments. For instance, for P1 (\fig~\ref{fig:divergence}), at turn 4, A1 assigned a rating of 5, while A2 scored it as 2, and A3 as 3. The participant's laughter posed a challenge as it was perceived both as a sign of high enjoyment (by A1) and, conversely, as an expression of frustration towards the situation or the robot (by A2 and A3), rather than amusement with the robot. Furthermore, at turn 15, A1 assigned a rating of 3, while A2 rated it 1, and A3 as 2. In this context, the participant remained entirely silent, awaiting the robot to initiate further interaction. Annotators interpreted this silence differently, seeing it as politeness, boredom, or discomfort.


For another participant (P25, \fig~\ref{fig:divergence}), during turn 8, A1 assigned a rating of 5, whereas A2 rated it as 1, and A3 as 2. The participant asked the robot to make more eye contact with them, which could be interpreted as a period of heightened immersion and anthropomorphism or criticism. Following this, the annotators consistently exhibited discord in their assessments until turn 17, when they reached a consensus once more. For instance, at turn 11, A1 marked it as a 5, A2 as 3, and A3 as 1. In this case, the participant expressed reservations about sharing personal information with the robot due to unfamiliarity, yet did so while smiling and posing a question to the robot in a playful tone. This complexity in the interaction exchange presented challenges for the annotators, as it encompassed a multitude of actions. While the verbal content suggested discomfort, the presence of laughter, smiling, and playful tonality indicated enjoyment. Consequently, the annotators encountered mixed signals, and the resulting ratings depended on which aspect of the interaction they prioritized.

The notable inconsistencies between A1's ratings and those of the others led to the inclusion of reliability results for the more consistently aligned group of annotators (A2 and A3) in \sect~\ref{results:reliability}.

\subsection{Similarities and Discrepancies Between User and Annotator Perceptions} 
\label{sec:discrepancies}

 \begin{figure}[b]
    \centering
    \includegraphics[width=0.93\columnwidth]{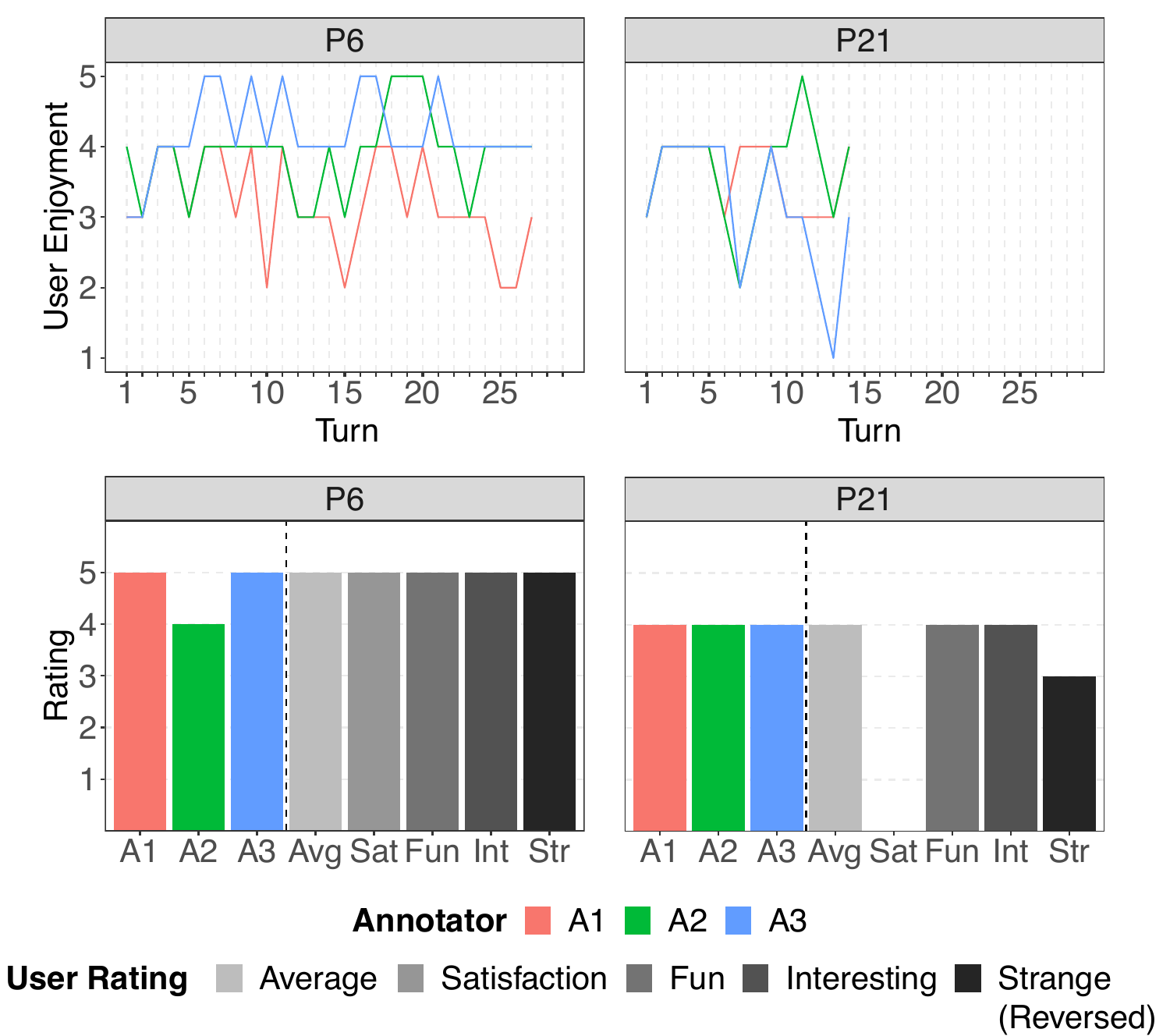}
    \caption{Similarity between annotator ratings and user perceptions (P6, P21).}
    \label{fig:similarity}
\end{figure}

While annotators aligned well with a large proportion of the participants in their perceptions, there were substantial discrepancies for some of the participants. For instance, for P6 and P21 (see \fig~\ref{fig:similarity}), it can be noted that the annotators and the participant interpreted user enjoyment in a similar way, with overall interaction scored as 5 and 4, matching that of the average of user reported values. The conversations went well, and the participants seemed to take a playful approach in the interaction, which was reflected in both the participants' and annotators' scores.

\begin{figure}[t]
    \centering
    \includegraphics[width=0.93\columnwidth]{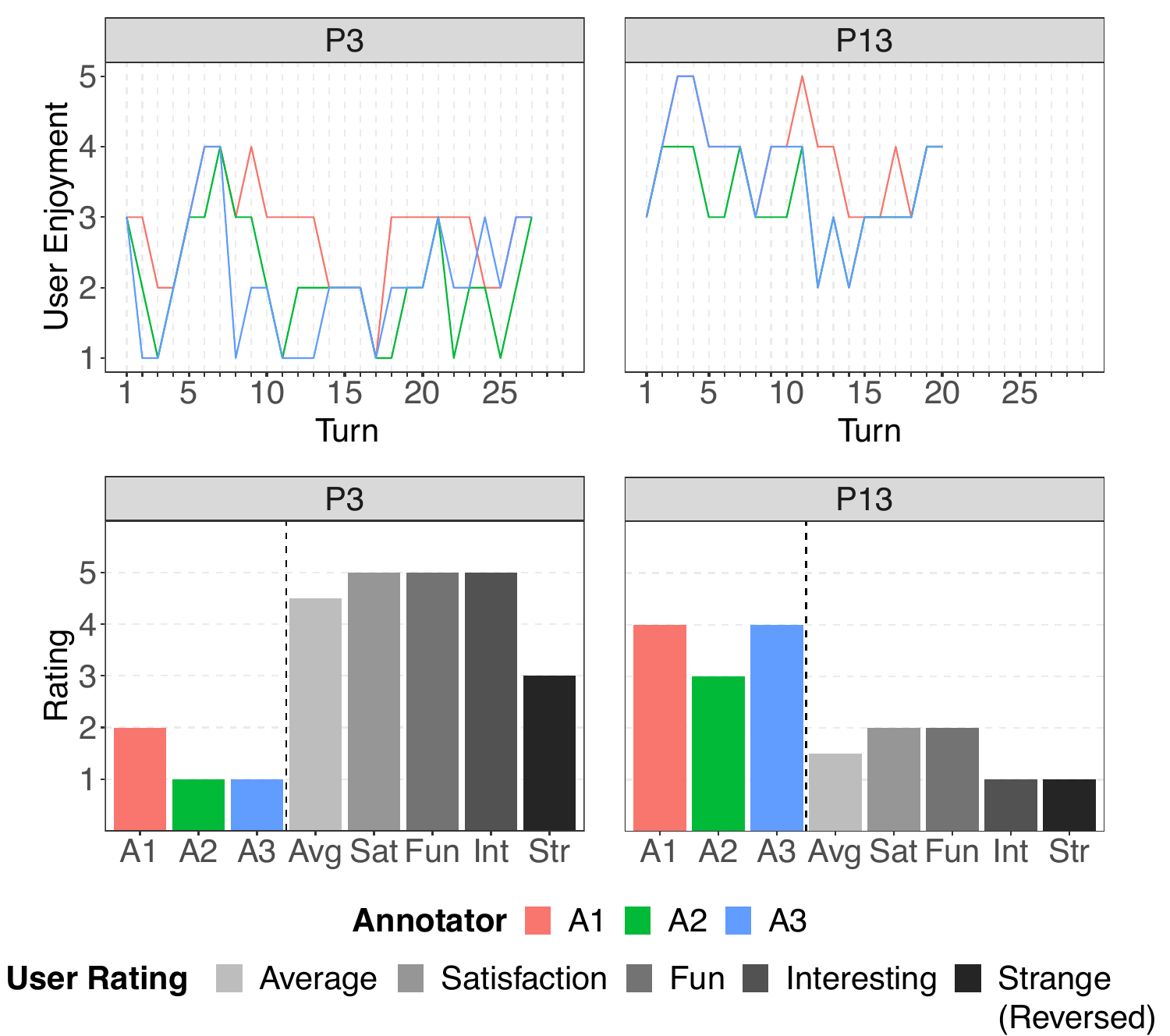}
    \caption{Discrepancy between annotator ratings and user perceptions (P3, P13).}
    \label{fig:discrepancy}
\end{figure}

The conversation context might cause a discrepancy between the annotators' assessment and users' rating of the enjoyment. For instance, P3 (\fig~\ref{fig:discrepancy}) talked about a controversial topic (UFOs) with the robot. The robot repeatedly questioned the participant (\eg \quotes{Why do you think that?}, \quotes{Can you tell more about what gave you this insight?}) when they were affirmative about having observed the existence of UFOs. The repetitive questions could have been perceived as offensive or discomforting due to the nature of the topic and their stance towards it, despite it being a type of interaction failure (repetition of the same phrase) that occurred with other participants as well. The participant stated to the robot that they wanted to change the topic twice, and then turned to the experimenters to voice this desire (after 3.5 minutes), in addition to displaying cues of anxiety (\eg playing with fingers, looking around at the camera and at the experimenters), which can confirm the belief from the annotators that the participant had a negative experience, who rated the interaction low in enjoyment. The participant managed to change the topic on their own to talk about the weather, and later about the university and robots for the rest of the conversation. However, the participant gave high scores (see \fig~\ref{fig:discrepancy}) in all aspects of enjoyment. 
The participant might have rated the experience as more enjoyable and interesting than the annotators due to researcher bias, \ie that the participant felt the need to please the researcher. Given the controversial topic discussed, the positive ratings can be interpreted as a strategy to avoid judgment from the researchers, as the participant frequently gazed at the experimenter during several exchanges, while displaying signs of enjoyment (\eg smiles, thumbs up, nods) even after the topic change. Another reason could be the novelty effect, since they might have been happy to talk with a robot regardless of the negative experience.

Another participant (P13, \fig~\ref{fig:discrepancy}) experienced the interaction as less enjoyable than the annotators interpreted. This might be because the participant was experiencing a high number of technical and social failures from the robot while still displaying enjoyable signs. The participant used the robot for transactional requests rather than a casual conversation. The transactional nature of the conversation combined with failures might explain why the participant gave low enjoyment scores. The annotators gave a higher score because the participant seemed to forgive the failures, laugh them off, and continue the conversation smoothly. This can be seen as an important reminder that the annotators are not always assessing the same aspects as the participants in their self-assessment.



\end{document}